\documentclass{article}
\usepackage[square, comma, sort&compress, numbers]{natbib}
\usepackage[final]{neurips_2024}

\usepackage[utf8]{inputenc} % allow utf-8 input
\usepackage[T1]{fontenc}    % use 8-bit T1 fonts
\usepackage{booktabs}       % professional-quality tables
\usepackage{amsfonts}       % blackboard math symbols
\usepackage{nicefrac}       % compact symbols for 1/2, etc.
\usepackage{microtype}      % microtypography
\usepackage{xcolor}         % colors
\usepackage{colortbl}
\usepackage[pagebackref=true,breaklinks,colorlinks,citecolor=blue,linkcolor=blue,urlcolor=blue,hypertexnames=false]{hyperref}
\usepackage{amsmath}
\usepackage{amssymb}
\usepackage{mathtools}
\usepackage{amsthm}

\theoremstyle{plain}

\theoremstyle{definition}

\theoremstyle{remark}

\usepackage{algorithm}
\usepackage{algorithmic}

\usepackage{hyperref}
\usepackage{graphicx}
\usepackage{amsmath}
\usepackage{amssymb}
\usepackage{makecell}
\usepackage{multirow}
\usepackage{overpic}
\usepackage{subcaption}
\usepackage[export]{adjustbox}
\usepackage{enumitem}
\usepackage{amsmath} 

 \newcommand\sotaa{\textcolor{red}}
\newcommand\sotab{\textcolor{blue}}
\def\vs{\emph{vs.\ }}
\def\eg{\emph{e.g.}}
\def\ie{\emph{i.e.}}

\def\etal{\emph{et al.\ }}

\title{Sharing Key Semantics in Transformer Makes Efficient Image Restoration}

\author{
Bin Ren$^{1,2,3}$\thanks{Work done during visiting at ETH Z\"urich and INSAIT Sofia University.} \quad
Yawei Li$^{4}$\thanks{Project Leader \& Corresponding Author. Email: \texttt{li.yawei.ai@gmail.com}}\quad
Jingyun Liang$^{4}$\quad
Rakesh Ranjan$^{5}$\quad
Mengyuan Liu$^{6}$\\
\textbf{Rita Cucchiara}$^{7}$\quad
\textbf{Luc Van Gool}$^{3}$\quad
\textbf{Ming-Hsuan Yang}$^{8}$\quad 
\textbf{Nicu Sebe}$^{2}$ \\
$^1$University of Pisa \quad
$^2$University of Trento \quad
$^3$INSAIT, Sofia University, \\
$^4$ETH Z\"urich \quad
$^5$Meta Reality Labs \quad
$^6$Peking University \\
$^7$University of Modena and Reggio Emilia \quad 
$^8$University of California, Merced\\
}

\begin{document}
\maketitle
\begin{abstract}
Image Restoration (IR), a classic low-level vision task, has witnessed significant advancements through deep models that effectively model global information. Notably, the emergence of Vision Transformers (ViTs) has further propelled these advancements. When computing, the self-attention mechanism, a cornerstone of ViTs, tends to encompass all global cues, even those from semantically unrelated objects or regions. This inclusivity introduces computational inefficiencies, particularly noticeable with high input resolution, as it requires processing irrelevant information, thereby impeding efficiency. Additionally, for IR, it is commonly noted that small segments of a degraded image, particularly those closely aligned semantically, provide particularly relevant information to aid in the restoration process, as they contribute essential contextual cues crucial for accurate reconstruction. 
To address these challenges, we propose boosting IR's performance by sharing the key semantics via Transformer for IR (\ie, SemanIR) in this paper. Specifically, SemanIR initially constructs a sparse yet comprehensive key-semantic dictionary within each transformer stage by establishing essential semantic connections for every degraded patch. Subsequently, this dictionary is shared across all subsequent transformer blocks within the same stage. This strategy optimizes attention calculation within each block by focusing exclusively on semantically related components stored in the key-semantic dictionary. As a result, attention calculation achieves linear computational complexity within each window. Extensive experiments across 6 IR tasks confirm the proposed SemanIR's state-of-the-art performance, quantitatively and qualitatively showcasing advancements. The visual results, code, and trained models are available at ~\url{https://github.com/Amazingren/SemanIR}.
\end{abstract}
%-----------------------%
\section{Introduction}
\label{sec:introduction}
%-----------------------%
Image restoration (IR) stands as a fundamental task within low-level computer vision, aiming to enhance the quality of images affected by numerous factors, including noise, blur, low resolution, compression artifacts, mosaic patterns, adverse weather conditions, and other forms of distortion. This capability holds broad utility across various domains, facilitating information recovery in medical imaging, surveillance, and satellite imagery. Furthermore, it bolsters downstream vision tasks like object detection, recognition, and tracking~\cite{sezan1982image,molina2001image,ren2024ninth}.
Despite notable progress in recent years, prevalent IR methods encounter challenges in effectively addressing complex distortions or preserving/recovering crucial image details~\cite{li2023efficient}. 
Achieving high-quality image recovery necessitates meticulous exploration of the rich information present in degraded counterparts.

\begin{figure}[!t]
    \centering
    \includegraphics[width=1.0\linewidth]{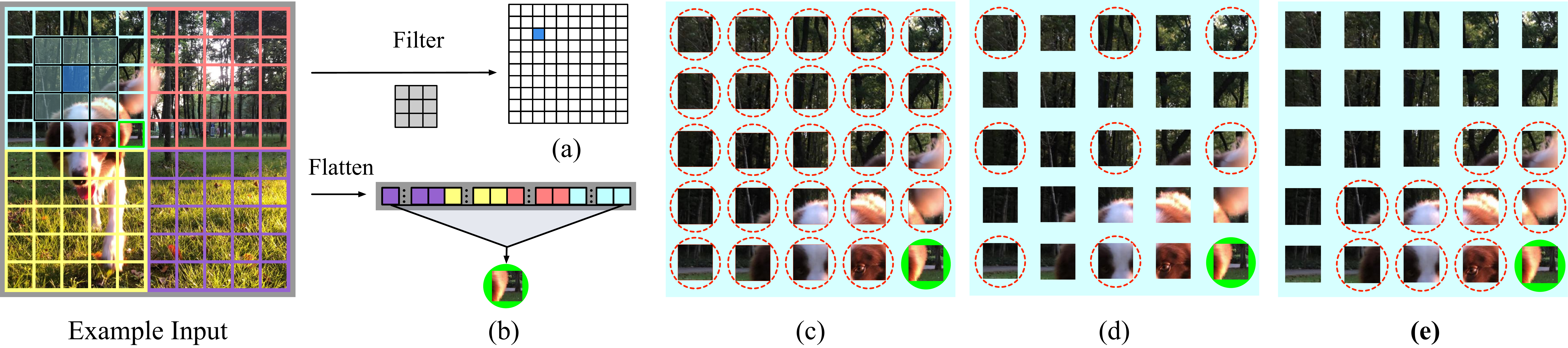}
    \caption{(a) The CNN filter captures information only within a local region. (b) The standard MLP/Transformer architectures take full input in a long-sequence manner. (c) The window-size multi-head self-attention (MSA) mechanism builds a full connection within each window. (d) Position-fixed sparse connection. (e) The proposed \textit{Key-Semantic} connection.}
    \label{fig:dog_demo}
    \vspace{-5mm}
\end{figure}

In modern IR systems, representative networks for learning rich image information are typically constructed using three fundamental architectural paradigms. \ie, convolutional neural networks (CNNs)~\cite{lecun1998gradient,zamir2021multi}, Multilayer perceptrons (MLPs)~\cite{bishop2006pattern,tu2022maxim}, and Vision Transformers (ViTs)~\cite{vaswani2017attention,dosovitskiy2020image}. The input image is treated as a regular grid of pixels in the Euclidean space for CNNs (Fig.~\ref{fig:dog_demo}(a)) or a sequence of patches for MLPs and ViTs (Fig.~\ref{fig:dog_demo}(b)). However, degraded inputs usually contain irregular and complex objects. While these choices perform admirably in scenarios with regular or well-organized object boundaries, they have limitations when applied to degraded images with more flexible and complex geometrical contexts.

Additionally, CNNs struggle to model long-range dependencies due to their limited receptive field (Fig.~\ref{fig:dog_demo}(a)). 
In contrast, MLPs and ViTs can capture long-range relations effectively, although at the cost of losing inductive bias and incurring a heavy computational burden, \ie, quadratic complexity increases with higher input resolution\cite{tu2022maxim,vaswani2017attention,dosovitskiy2020image,ren2023masked}. 
To address these limitations, recent IR methods have explored strategies for complexity reduction. 
A common approach is to implement MSA within local image regions~\cite{liu2021swin,liang2021swinir}. 
For example, SwinIR~\cite{liang2021swinir} and GRL~\cite{li2023efficient} employ full MSA or region-fixed anchored stripe MSA, but they still struggle with irregular object connections. Furthermore, prior research~\cite{zontak2011internal} highlights that smooth image content is more prevalent than complex image details, underscoring the need for distinct treatment based on semantic content.

In this paper, we introduce a novel approach, SemanIR, to address the limitations above. Specifically, within each transformer stage, we first construct a key-semantic dictionary, which stores only the top-k semantically related relations for each given degraded patch with the k-nearest neighbors (\ie, KNN) algorithm. Then, the attention operation within each transformer layer occurs only among the top-k patches. This design brings two main advantages, \ie, 1) Each degraded patch benefits from its semantically similar patches, typically containing comparable contextual or textural information, while excluding the side effects from other patches that contain entirely unrelated information. 2) Compared to the conventional window-wise attention, which built a dense connection between all the patches (Fig.~\ref{fig:dog_demo}(c)) that leads to highly computationally demanding, or 
a sparse but position-fixed manner (Fig.~\ref{fig:dog_demo}(d)) which introduces irrelevant semantics. Our key-semantic connection (Fig.~\ref{fig:dog_demo}(e)) leads to a sparse yet more meaningful attention operation, which allows our method to achieve the same receptive field as previous ViTs-based methods while maintaining lower computational costs. 
This is not like previous token merging or pruning methods~\cite{bolya2022tome,yin2022vit,rao2021dynamicvit} that may merge unrelated information or prune some semantically related information. In addition, to make the proposed method more efficient, instead of creating a key-semantic dictionary for each transformer layer, we create it just once at the beginning of each transformer stage and then share it with all the following transformer layers within the same stage. 
This not only largely reduced the computation burden but also made our methods different from other token pruning and merging methods~\cite{zhang2024vision,xia2022vision,zhang2024you}, which include dynamic patch skipping/selection within each attention layer or involve an additional offset generation network. Meanwhile, merging or pruning tokens will lead to a loss of information in corresponding patches, which is not preferred in image restoration~\cite{venkataramananskip}. In addition, such a sharing strategy allows each degraded patch to be continuously optimized by its semantically related patches within each stage.

It is also worth noting that the implementation of the attention layer of our SemanIR is achieved in three interesting manners (\ie, Triton~\cite{dao2022flashattention}\footnote{Open-source GPU programming tool \url{https://openai.com/research/triton}.}, torch-mask, and torch-gather), which are discussed in our ablation studies. Overall, our method’s suitability for image restoration comes from the utilization of semantic information, the preservation of the details, and the effective KNN strategy. 
The contributions of this work are:
\begin{enumerate}[nosep]
    \item For each degraded input patch, we propose to construct a key-semantic dictionary that stores its most semantically relevant $k$ patches in a sparse yet representative manner. This strategy excludes the side effects of a given degraded patch from semantically unrelated parts.

    \item Based on the constructed key-semantic dictionary, we propose to share the key semantic information across all the attention layers within each transformer stage, which not only makes each degraded patch well-optimized but also largely reduces the computational complexity compared to conventional attention operations.
    
    \item Extensive experimental results show that the proposed SemanIR achieves state-of-the-art performance on 6 IR tasks, \ie, deblurring, JPEG compression artifact removal (JPEG CAR), denoising, IR in adverse weather conditions (AWC), demosaicking, and classic image super-resolution (SR).
\end{enumerate}
%-----------------------%
\section{Related Work}
\label{sec:related-work}
%-----------------------%
\noindent{\textbf{Image Restoration (IR),}} as a long-standing ill-posed inverse problem, is designed to reconstruct the high-quality image from the corresponding degraded counterpart with  
numerous applications~\cite{richardson1972bayesian,banham1997digital,li2023lsdir}. 
Initially, IR was addressed through model-based solutions, involving the search for solutions to specific formulations.
However, learning-based approaches have gained much attention with the significant advancements in deep neural networks. 
Numerous approaches have been developed, including regression-based~\cite{lim2017enhanced,lai2017deep,liang2021swinir,li2023efficient,zhang2024transcending} and generative model-based pipelines~\cite{gao2023implicit,wang2022zero,luo2023image,yue2023resshift}.   
In this paper, we propose a regression-based method for image restoration. 

\noindent{\textbf{Non-Local Priors Modeling in IR.}} Tradition model-based IR methods reconstruct the image by regularizing the results (\emph{e.g.}, Tikhonov regularization~\cite{golub1999tikhonov}) with formulaic prior knowledge of natural image distribution. However, it is challenging for these methods to recover realistic detailed results with hand-designed priors. 
Besides, some other classic method finds that self-similarity is an effective prior, which leads to an impressive performance~\cite{buades2005non,dabov2007image}. 
Apart from traditional methods, the non-local prior has also been utilized in modern deep learning networks~\cite{wang2018non,li2023efficient,zhang2019residual}, typically captured by the self-attention mechanism. 
More recently, the overwhelming success of transformers~\cite{vaswani2017attention} in the natural language processing domain~\cite{khan2022transformers} and the classic vision community~\cite{dosovitskiy2020image,carion2020end,touvron2021training,ye2019cross,chen2021pre} has led to the development of numerous ViT-based IR methods. These methods aim to enhance the learning ability for modeling non-local priors~\cite{liang2021swinir,zamir2022restormer,li2023efficient,chen2023activating,chen2022cross,zhang2023accurate} and consequently archives better performance. Meanwhile, this raises a question: are all non-local priors essential for IR?

\noindent{\textbf{Key-Semantic Non-local Prior Exploration for IR.}} To answer the question, we found many methods demonstrating the effectiveness of modeling key semantics within ViTs. For example, KiT~\cite{lee2022knn} proposed increasing the non-local connectivity between patches at different positions through KNN matching. This approach aims to better capture the non-local relations between the base patch and other patches in each attention calculation. However, it results in significant additional computational costs due to the KNN matching. DRSformer~\cite{chen2023learning} proposed a top-k selection strategy that chooses the most relevant tokens to model the non-local priors for draining after each self-attention calculation without reducing the computation complexity, since after each attention calculation, the DRSFormer utilized (mask, top-k, scatter) operations at each transformer layer. Consequently, this inevitably increases the computation cost. Similar conclusions can be also drawn from the graph perspective solutions~\cite{gori2005new,scarselli2008graph,mou2021dynamic,jiang2023graph} for various IR tasks, like facial expression restoration~\cite{liu2020facial}, image denoising~\cite{simonovsky2017dynamic}, and artifact reduction~\cite{mou2021dynamic}. \cite{jiang2023graph} construct the graph with transformer-based architecture where each patch is connected to all other patches. All these methods suggest that if the semantically related information can be addressed, the degraded image can be restored with better performance. 
However, the efficiency issue, which is extremely unignorable, remains untouched within the aforementioned methods. 
It is particularly crucial for ViTs-based image restoration methods, which often need to address high-resolution degraded input images. 
LaViT~\cite{zhang2024you} reduces computational costs by storing attention scores from a few initial layers and reusing them in subsequent layers. 
However, this approach does not change the computation cost of attention itself; it merely reuses previously computed scores. In this paper, we propose sharing key semantics within each transformer stage, demonstrating its efficiency and effectiveness through experimental and theoretical analysis. Our method, SemanIR, reduces computation in both training and inference by using a semantic dictionary to filter out irrelevant patches during training and optimizing attention operations with Triton kernels during inference.

\begin{figure*}[!t]
    \centering
    \includegraphics[width=1.0\linewidth]{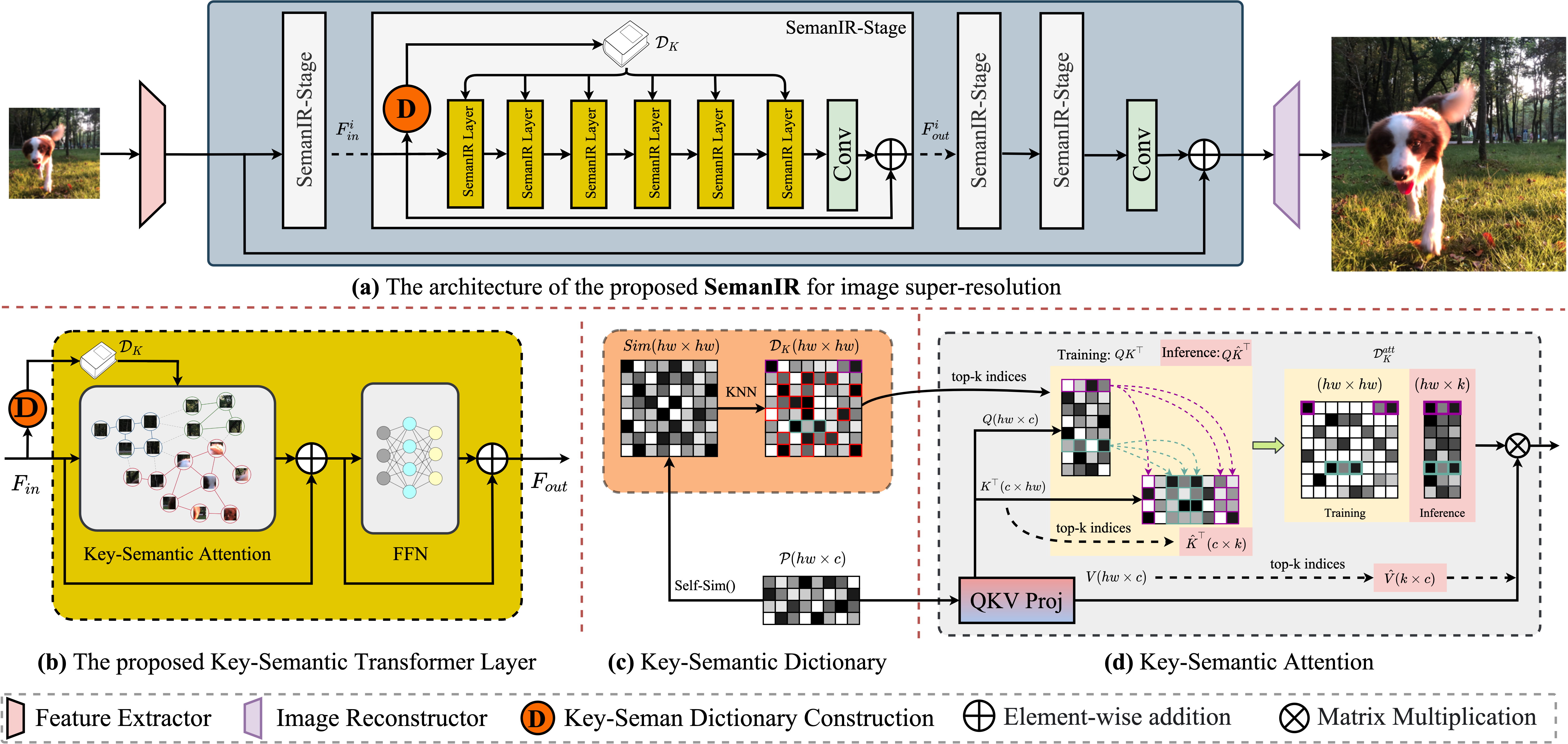}
    \caption{The proposed SemanIR mainly consists of a convolutional feature extractor, the main body of SemanIR for representation learning, and an image reconstructor. 
    The main body in columnar shape shown here is for image SR, while the U-shaped structure (shown in \textit{Appx.}~\ref{subsec:appx_architecture}) is used for other IR tasks. (b) The transformer layer of our SemanIR. The toy example of $k$=3 for (c) the Key-semantic dictionary construction and (d) the attention of each Layer.}
    \vspace{-1em}
    \label{fig:framework}
\end{figure*}
%-----------------------%
\section{Methodology}
\label{sec:method}
%-----------------------%
To comprehensively study the effectiveness of the proposed method that is architecture-agnostic for various IR tasks, we adopted two of the most commonly used architectures \ie the columnar architecture (shown in Fig.~\ref{fig:framework} (a)) for image SR and the U-shaped architecture (shown in the \textit{Appendix}, \ie, \textit{Appx.}~\ref{subsec:appx_architecture}) for other IR tasks. 
In the following, we first show how to construct the key-semantic dictionary in Sec.~\ref{subsec:key_semantic_sharing}. 
Based on the key-semantic dictionary, then we explain why sharing it works for IR, and we introduce the basic unit, the key-semantic transformer layer in Sec.~\ref{subsec:key_semantic_layer}. 
Finally, two interesting discussions (Sec.\ref{subsec:discuss}) are introduced regarding the implementation style of the Key-Graph attention and two top-k settings during the training. The efficiency analysis is provided in \textit{Appx.}~\ref{subsec:appx_efficiency}.

\subsection{Key-Semantic Dictionary Construction}
\label{subsec:key_semantic_sharing}
Consider the input feature $F_{in} \in \mathbb{R}^{H \times W \times C}$, where $H$, $W$, and $C$ denote the height, the width, and the channel. ViTs are good at modeling global dependencies for $F_{in}$. This is achieved by the MSA, the core of ViTs, by connecting all other patches to a certain patch. 
Specifically, $F_{in}$ is first split into $N$ non-overlapping patches, forming its patch representation $\mathcal{P} = \{p_{i}| p_{i} \in \mathbb{R}^{ hw \times c}, i = 1, 2, 3, ..., N\}$, where $h$, $w$, and $c$ are the height, the width, and the channel of each patch. To achieve such global connectivity  $\mathcal{D}$, $\mathcal{P}$ is linearly projected
into Query ($Q$), Key ($K$), and Value ($V$) matrices, which are denoted as $Q=\mathcal{P}\mathbf{W}_{qry}$, $K=\mathcal{P}\mathbf{W}_{key}$, and $V=\mathcal{P}\mathbf{W}_{val}$. $\mathbf{W}_{qry/key/val}$ represents the learnable projection weights. Then $\mathcal{D}$ is performed by a softmax function as follows:
\begin{equation}
    \mathcal{D}_{ij} = \frac{\operatorname{exp}(Q_{i}K_{j}^{\top})}{\sum_{k = 1...j} \operatorname{exp}(Q_{i}K_{k}^{\top}/\sqrt{d})}, i = 1, 2, 3, ..., N, 
    \label{eq:eq1}
\end{equation}
where $d$ is the dimension of $Q$ and $K$. Then each patch is aggregated via $\sum_i\mathcal{D}_{ij} V_{i}$. 
However, $\mathcal{D}_{ij}$ functions as a full semantic dictionary, where each patch is connected to all other patches regardless of their semantic relatedness. \eg, given a sub-graph with a green dog patch shown in Fig.~\ref{fig:dog_demo}(c), the tree-related patches are also considered. 
Since such an operation occurs at each attention calculation step in ViTs, it inevitably increases the computational cost, especially for large-scale inputs. 
In addition, for IR, a degraded patch usually benefits from its most semantically related patches, as they share similar texture and geometric information. This naturally raises the question: \textit{Can we build a key-semantic dictionary, $\mathcal{D}_{K}$, where each patch is connected only to its most related patches?}

To mitigate this problem, given $\mathcal{P}$, we first construct a fully connected dictionary $\mathcal{D}$ by calculating its self-similarity $\operatorname{Sim()}$ via a naive dot product operation as $\mathcal{D}(i, j) = \operatorname{Sim}(i, j ) = p_i \cdot p_j^{\top}$, which describes the correlation among all the patches, with higher values indicating stronger correlations.
% In this context, each destination patch $p_i$ is connected to all other patches $p_j$, irrespective of the degree of semantic relatedness between $p_i$ and $p_j$.
To reduce the side influence of patches with low correlation (\eg, the tree-related patches at the upper left part in Fig.~\ref{fig:dog_demo} (c)) for the green background dog destination patch, we keep only $k$ highly related patches and exclude the remaining.
This is achieved by a KNN algorithm from $\mathcal{D}$ as follows:
\begin{equation}
\mathcal{D}_{K}(i, j) = 
    \begin{cases}
    \mathcal{D}(i, j),~\mathcal{D}(i, j) \geq \operatorname{Sim}(i, )_{k} ~\text{and}~ i \neq j
    \\ 
    0,~~~~~~~~~~\text{otherwise},
    \end{cases}
\end{equation}
where $\operatorname{Sim}(i, )_{k}$ denotes the $k_{th}$ largest connectivity value of patch $p_{i}$.
As a result, $\mathcal{D}_{K}$ contains only the patches with high correlation (\eg, dog-related patches in Fig.~\ref{fig:dog_demo}(e)) for the destination patch (\eg, the green dog patch).
We formalize the key-semantic dictionary construction as $\operatorname{KeySemanDictionary\_Constructor}()$ in Alg.~\ref{alg:semanir_stage}.
Although such a dictionary allows the subsequent attention operation to focus on the most semantically related patches, constructing it before each attention operation significantly increases computational costs.
Meanwhile, we observed that IR architectures typically use transformers stage-by-stage (See Fig.~\ref{fig:framework}(a)). 
This means that in each stage, several transformer layers are directly connected sequentially and operate at the same semantic level. Inspired by this, we propose to \textit{share the same key-semantic dictionary} for all the transformer layers.

\begin{algorithm}[!t]
    \small
    \renewcommand{\algorithmicrequire}{\textbf{Input:}}
    \renewcommand{\algorithmicensure}{\textbf{Output:}}
    \caption{Key-Semantic Transformer Stage (\ie, SemanIR Stage)}
    \label{alg:semanir_stage}
    \begin{algorithmic}[1]
        \REQUIRE input feature ${F_{in}}$, numbers of SemanIR layer $N_{layer}$, KNN value $k$, the patched feature   {$\mathcal{P}$}
	\ENSURE aggregated feature $F_{out}$
            \STATE $\mathcal{D}_{K} \leftarrow \operatorname{KeySemanDictionary\_Constructor}( {\mathcal{P}}, k)$ // Sec.~\ref{subsec:key_semantic_sharing}
            \FOR{$i=1$ to  {$N_{layer}$ }}
                \STATE $Q, K, V \leftarrow \operatorname{Linear\_Proj}( {\mathcal{P}})$ 
                \STATE $\hat{ {\mathcal{P}}} \leftarrow 
                \operatorname{SemanIR\_Att}(Q, K,  {V}, \mathcal{D}_{K})$ // Sec.~\ref{subsec:key_semantic_layer}
                \STATE $ {\mathcal{P}} \leftarrow \hat{ {\mathcal{P}}} + \operatorname{FFN}(\hat{ {\mathcal{P}}})$ // Sec.~\ref{subsec:key_semantic_layer}
            \ENDFOR
            \STATE $F_{out} \leftarrow F_{in} + \operatorname{Conv}( {\operatorname{UnPatch}(\mathcal{P})})$ 
		\STATE \textbf{return} $F_{out}$
    \end{algorithmic}  
\end{algorithm}

\subsection{Sharing Key Semantics Cross Transformer Layers}
\label{subsec:key_semantic_layer}
The structure of each transformer layer is shown in Fig.~\ref{fig:framework}(b), which consists of a key-semantic attention block followed by a feed-forward network (FFN). 
Specifically, given an input $F_{in}$, we form each transformer layer as $z = \operatorname{FFN}(f_{\theta}(F_{in}))$, where $f$ means the transformer layer, $z$ denotes the output, and $\theta$ is the trainable parameters. 
Previous methods tried to reduce the computation cost mainly by applying some techniques (\ie, $\mathcal{T}$) like token merging or pruning after each attention calculation or the entire transformer layer, and it can be formalized as $z = \mathcal{T}(\operatorname{FFN}(f_{\theta}(F_{in})))$ or $z = (\operatorname{FFN}(\mathcal{T}f_{\theta}(F_{in})))$. However, the main computation cost from MSA is still untouched.

Owing to the permutation-invariant property (\ie, $f_{\theta}(\mathcal{T}x) = \mathcal{T}f_{\theta}(x)$, here $\mathcal{T} \in \mathbb{R}^{N \times N}$ means any token level permutation matrix) inherent in both the MSA and the FFN~\cite{vaswani2017attention,lee2019set}, the transformer layer consistently produces identical representations for patches that share the same attributes, regardless of their positions or the surrounding structures~\cite{chen2022structure}. 
In other words, patches at the same location are consistently connected to other patches possessing the same attributes as they traverse through the various layers within the same stage. 
It enables $\mathcal{D}_{K}$ to serve as a reference permutation for each attention in the subsequent transformer layers, facilitating efficient yet highly semantics-related attention operations. 
This distinguishes our method from previous token merging/pruning~\cite{zhang2024vision,xia2022vision} or sparse attention-based methods (Fig.\ref{fig:dog_demo}(d)) that only activate patches in a grid-fixed manner\cite{zhang2023accurate}.

The workflow is intuitively illustrated in Fig.~\ref{fig:framework} (c) and (d). 
Initially, the patch $\mathcal{P}$ is linear projected via $\operatorname{Linear\_Proj()}$ (The 3rd step in Alg.~\ref{alg:semanir_stage}) into $Q$, $K$, and $V$.
For each patch $p_i$ in $Q$, instead of calculating the self-attention with all $hw$ patches in $K$\&$V$, only $k$ essential patches are selected via the semantic lookup via the indices from $\mathcal{D}_{K}$ in them, forming the $\hat{K}$\&$\hat{V}$. 
Then the attention matrix $\mathcal{D}_{K}^{att}$ is obtained by: $\mathcal{D}_{K}^{att} = \operatorname{Softmax_{K}}(Q\hat{K}^{\top}/ \sqrt{d})$, which captures the pair-wise relation between each destination patch $p_i$ in $Q$ with only $k$ patches in $K$\&$V$ that are semantically highly related to $p_i$. 
For other unselected patches in $K$\&$V$, we aim to maintain their position in their corresponding places without any computation. Based on $\mathcal{D}_{K}^{att}$, the attention outputs the updated feature $\hat{\mathcal{P}}$ via: $\hat{\mathcal{P}} = \mathcal{D}_{K}^{att} \hat{V}$. We formulate these two procedures as $\operatorname{SemanIR\_Att()}$ in the 4th step of Alg.~\ref{alg:semanir_stage}. 
This differs from the conventional MSA, which calculates the relation of each patch in $Q$ and all patches in $K$\&$V$. Finally, with FFN, the output of each transformer layer is achieved via the 5th step of Alg.~\ref{alg:semanir_stage}.

Conversely, our design offers two advantages. Firstly, the computational cost can be significantly reduced within each attention window (detailed analysis can be found in the \textit{Appx.}~\ref{subsec:appx_efficiency}), enhancing efficiency. 
Additionally, sharing the key semantics across transformer layers within each stage acts as a loop that continuously optimizes a degraded patch with its most semantically related patches, ensuring the performance of the proposed method (supported by our experimental results in Sec.~\ref{sec:experiments}).

\subsection{Discussion}
\label{subsec:discuss}
\noindent \textbf{Fixed top-k \vs Random top-k Training Strategies.} 
In the fixed top-k approach, $k$ remains constant at 512 during training. 
In contrast, in the random top-k method, $k$ is randomly selected from the set $[64, 128, 192, 256, 384, 512]$. It is important to note that even in the random top-k setting, a fixed k value is maintained for all patches/pixels in each iteration.
During inference, the random top-k strategy offers more flexibility and requires training only a single model, making it more user-friendly and less resource-intensive.

\noindent \textbf{Implementation of the Attention of SemanIR.} 
To achieve the attention operation of the proposed SemanIR, we explored three different manners for the implementation, \ie, 
(i) \textit{Triton}, 
(ii) \textit{Torch-Gather}, and 
(iii) \textit{Torch-Mask}. 
Specifically, (i) is based on FlashAttention~\cite{dao2022flashattention}, and a customized GPU kernel is written for the operators proposed in this paper. Parallel GPU kernels are called for the nodes during run time. 
(ii) means that we use the `torch.gather()' function in PyTorch to choose the corresponding $Q_{gather}$ and $K_{gather}$ based on $\mathcal{D}_{K}$, then the attention operation is conducted between $Q_{gather}$ and $K_{gather}$. 
(iii) denotes that we keep only the value of selected patches of $\mathcal{D}_{K}$ and omitting other patches with low correlation via assigning those values to $-\infty$ guided by $\mathcal{D}_{K}$. 
Discussions of the pros and cons regarding these manners are provided in Sec.~\ref{subsec:ablatin}.

%-----------------------%
\section{Experiments}
\label{sec:experiments}
%-----------------------%
\begin{table*}[!t]
\parbox{.43\linewidth}{
\scriptsize
\begin{center}
\caption{GPU memory footprint of different implementations (\ie, Triton, Torch-Gather, and Torch-Mask) of our key-graph attention block. $N$ is the number of tokens and $k$ is the number of nearest neighbors. OOM denotes "out of memory".}
\label{table:ablation_implementation}
\vspace{0.9mm}
\setlength{\extrarowheight}{0.9pt}
\setlength{\tabcolsep}{7pt}
% \scalebox{0.7}{
\begin{tabular}{c|ccc}
\toprule[0.15em]

$N$	&Triton	&Torch-Gather	&Torch-Mask	\\ \midrule
512	&0.27 GB	&0.66 GB	&0.36 GB	\\
1024	&0.33 GB	&1.10 GB	&0.67 GB	\\
2048	&0.68 GB	&2.08 GB	&1.91 GB	\\
4096	&2.61 GB	&4.41 GB	&6.83 GB	\\
8192	&10.21 GB	&10.57 GB	&26.42 GB	\\ \midrule[0.15em]

$k$	&Triton	&Torch-Gather	&Torch-Mask	\\ \midrule
32	&5.51 GB	&15.00 GB	&13.68 GB	\\
64	&5.82 GB	&27.56 GB	&13.93 GB	\\
128	&6.45 GB	&OOM	&14.43 GB	\\
256	&7.70 GB	&OOM	&15.43 GB	\\
512	&10.20 GB	&OOM	&17.43 GB	\\

\bottomrule[0.15em]
\end{tabular}
% }
\end{center}
}
\hfill
\parbox{.55\linewidth}{\scriptsize\centering
\caption{\textit{\textbf{Single-image motion deblurring}} results. {GoPro}~\cite{nah2017deep} dataset is used for training. 
% Top-2 results are highlighted in \textcolor{red}{red} and \textcolor{blue}{blue}, respectively. 
}
\label{table:motion_deblurring}
\vspace{-2.5mm}
\setlength{\extrarowheight}{0.1pt}
\setlength{\tabcolsep}{1pt}
% \scalebox{0.69}{
\begin{tabular}{l | cc | cc | cc}
\toprule[0.15em]
 & \multicolumn{2}{c|}{\textbf{GoPro} } & \multicolumn{2}{c|}{\textbf{HIDE} } & \multicolumn{2}{c}{Average} \\
 \textbf{Method} & PSNR$\uparrow$ & {SSIM$\uparrow$} & PSNR$\uparrow$ & {SSIM$\uparrow$} & PSNR$\uparrow$ & {SSIM$\uparrow$} \\
\midrule[0.1em]
DeblurGAN~\cite{deblurgan}	&28.70 & 0.858		&24.51 & 0.871		&26.61 & 0.865		\\
Nah~\etal~\cite{nah2017deep}	&29.08 & 0.914		&25.73 & 0.874		&27.41 & 0.894		\\
% Zhang~\etal~\cite{zhang2018dynamic}	&29.19 & 0.931		&- & -		&- & -		\\
DeblurGAN-v2~\cite{deblurganv2}	&29.55 & 0.934		&26.61 & 0.875		&28.08 & 0.905		\\
SRN~\cite{tao2018scale}	&30.26 & 0.934		&28.36 & 0.915		&29.31 & 0.925		\\
% Shen \etal \cite{shen2019human}	&- & -		&28.89 & 0.930		&- & -		\\
Gao \etal \cite{gao2019dynamic}	&30.90 & 0.935		&29.11 & 0.913		&30.01 & 0.924		\\
DBGAN \cite{zhang2020dbgan}	&31.10 & 0.942		&28.94 & 0.915		&30.02 & 0.929		\\
MT-RNN \cite{mtrnn2020}	&31.15 & 0.945		&29.15 & 0.918		&30.15 & 0.932		\\
DMPHN \cite{dmphn2019}	&31.20 & 0.940		&29.09 & 0.924		&30.15 & 0.932		\\
Suin \etal \cite{Maitreya2020}	&31.85 & 0.948		&29.98 & 0.930		&30.92 & 0.939		\\
CODE~\cite{zhao2023comprehensive} & 31.94 & - & 29.67 & - & 30.81  &  -  \\
SPAIR~\cite{purohit2021spatially_spair}	&32.06 & 0.953		&30.29 & 0.931		&31.18 & 0.942		\\
MIMO-UNet+~\cite{cho2021rethinking_mimo}	&32.45 & 0.957		&29.99 & 0.930		&31.22 & 0.944		\\
IPT~\cite{chen2021pre}	&32.52 & -		&- & -		&- & -		\\
MPRNet~\cite{zamir2021multi}	&32.66 & 0.959		&30.96 & 0.939		&31.81 & 0.949		\\
KiT~\cite{lee2022knn} &  32.70 & 0.959 & 30.98 & \sotaa{0.942}  & 31.84 & 0.951  \\  
NAFNet~\cite{chen2022simple} & 32.85 & 0.960 & - & - & - & - \\
Restormer~\cite{zamir2022restormer}	&32.92 & 0.961		&\textcolor{red}{31.22} & \textcolor{red}{0.942}	&32.07 & \textcolor{blue}{0.952}
		\\
Ren \etal~\cite{ren2023multiscale} & \sotab{33.20} & \sotab{0.963} & 30.96 & 0.938 & \sotab{32.08} & 0.951  \\
SemanIR (ours)	&\textcolor{red}{33.44} & \textcolor{red}{0.964}		&\textcolor{blue}{31.05} & \textcolor{blue}{0.941}		&\textcolor{red}{32.25} & \textcolor{red}{0.953}	\\
\bottomrule[0.15em]
\end{tabular}
}
\vspace{-7mm}
\end{table*}
\begin{figure}[!t]
    \centering
    \includegraphics[width=1\linewidth]{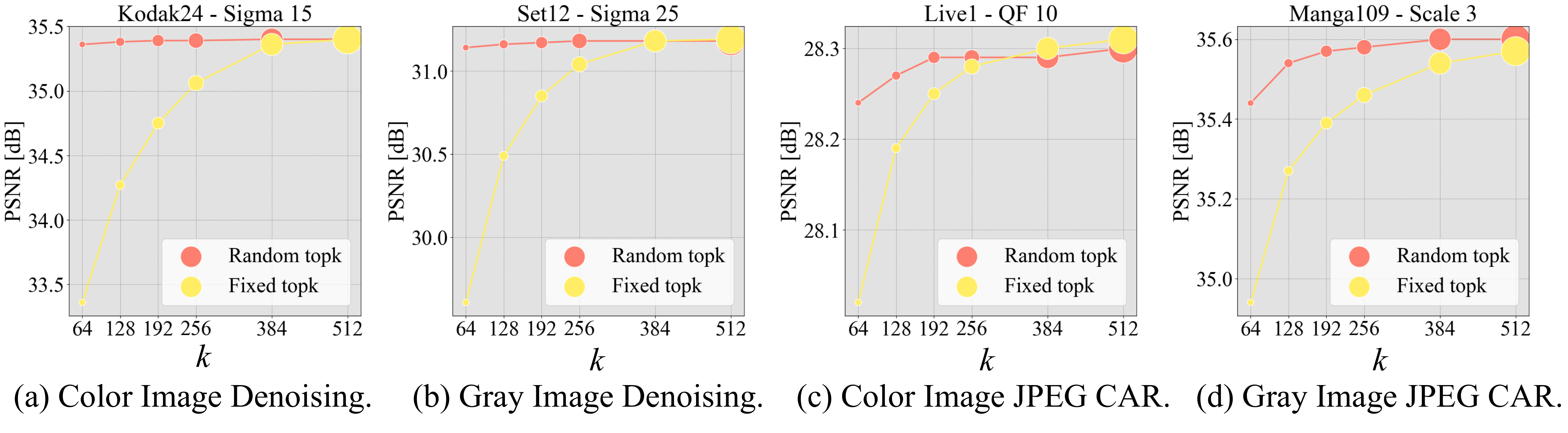}
    \vspace{-6.5mm}
    \caption{The impact of $k$ with different inference $k$ value. Circle size represents FLOPs.}
    \label{fig:main_figure_K}
    \vspace{-3.5mm}
\end{figure}
In this section, we first analyze three important ablation studies of our SemanIR, followed by extensive experiments on \textbf{6} IR tasks, \ie, deblurring, JPEG CAR, image denoising, IR in AWC, image demosaicking, and image SR.  More details about the architecture design, training protocols, the training/testing dataset, and full quantitative/additional qualitative results are shown in \textit{Appx.}~\ref{sec:appx_experimental} to ~\ref{sec:appx_results}.
The best and the 2nd-best results are reported in \textcolor{red}{red} and \textcolor{blue}{blue}, respectively. 
Note that \textcolor{magenta}{\textdagger} denotes a single model that is trained to handle multiple degradation levels \ie, noise levels, and quality factors.

\subsection{Ablation Study}
\label{subsec:ablatin}
\noindent\textbf{The impact of the implementation of SemanIR Attention} is assessed in terms of 
(i) \textit{Triton}, 
(ii) \textit{Torch-Gather}, and 
(iii) \textit{Torch-Mask} under different numbers of N (various from 512 to 8192) and K (various from 32 to 512). The results of the GPU memory footprint are shown in Tab.~\ref{table:ablation_implementation}, which indicate that \textit{Torch-Gather} brings no redundant computation while requiring a large memory footprint. 
Though \textit{Torch-Mask} brings the GPU memory increase, the increment is affordable compared to \textit{Torch-Gather} and also easy to implement. 
\textit{Triton} largely saves the GPU memory while at the cost of slow inference and difficult implementation for the back-propagation process. 
To optimize the efficiency of our SemanIR, we recommend employing \textit{Torch-Mask} during training and \textit{Triton} during inference, striking a balance between the efficiency and the GPU memory requirement.
\begin{figure}[!t]
    \centering
    \includegraphics[width=1\linewidth]{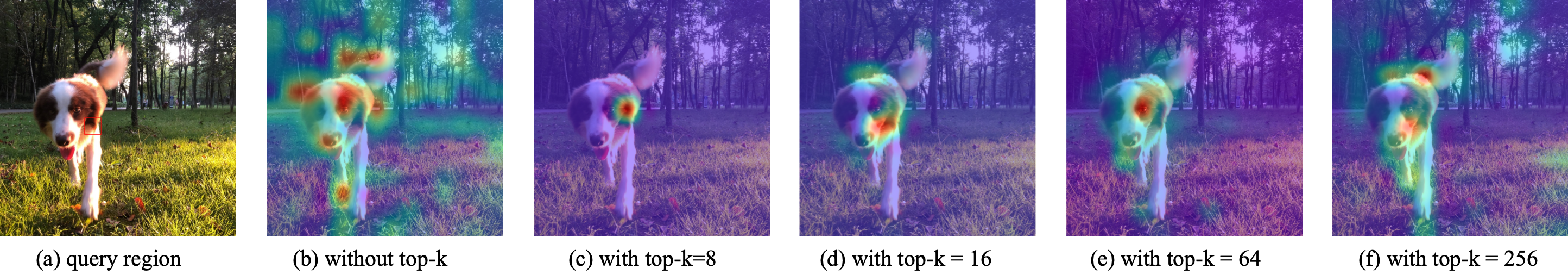}
    \vspace{-5mm}
    \caption{The impact of $k$ with different inference $k$ value.}
    \label{fig:dogvis_topk}
    \vspace{-5mm}
\end{figure}
\begin{table}[!t]
    \centering
    \caption{The \textbf{efficiency comparisons} results on Urban100 dataset.}
    \label{table:ab_efficiency}
    \vspace{1mm}
    \setlength{\tabcolsep}{6.0pt}
    \setlength{\extrarowheight}{0.1pt}
    \scalebox{0.77}{
    \begin{tabular}{c|c|c|cccc}
    \toprule[0.15em]
    \textbf{Task}	&\textbf{Method}	&\textbf{Architecture}	&\textbf{Params [M]}$\downarrow$	&\textbf{FLOPs [G]}$\downarrow$	&\textbf{Runtime [ms]}$\downarrow$	&\textbf{PSNR}$\uparrow$	\\ \hline
    \multirow{4}{*}{$\times4$ SR}	&SwinIR~\cite{liang2021swinir}	&Columnar	&11.90	&215.32	&152.24	&27.45	\\
    	&CAT~\cite{chen2022cross}	&Columnar	&16.60	&387.86	&357.97	&27.89	\\
    	&HAT~\cite{chen2023activating}	&Columnar	&20.77	&416.90	&368.61	&28.37	\\ 
    	&SemanIR-S (Ours)	&Columnar	&12.02	&290.20	&211.94	&28.34	\\ \midrule
    \multirow{4}{*}{\makecell{Denoising ($\sigma=50$) \\ (The same architecture \\ for other IR task)}}	&SwinIR~\cite{liang2021swinir}	&Columnar	&11.75	&752.06	&1772.84	&27.98	\\
    	&Restormer~\cite{zamir2022restormer}	&U-shape	&26.10	&154.88	&210.44	&28.29	\\
    	&GRL~\cite{li2023efficient}	&Columnar	&19.81	&1361.77	&3944.17	&28.59	\\
    	&SemanIR (Ours)	&U-shape	&25.85	&135.26	&240.05	&28.63	\\ \bottomrule[0.15em]
    \end{tabular}
    }
    \vspace{-5mm}
    % }
\end{table}

\noindent\textbf{The Impact of the $k$ in Key-Semantic Dictionary Construction.} 
Three interesting phenomena are observed from the results shown in Fig.~\ref{fig:main_figure_K} regarding the two top-k training strategies (Sec.~\ref{subsec:discuss}).
(1) The PSNR can largely increase with the increase of $k$ in a fixed manner. 
(2) When $k$ reaches a certain number (\ie, 384), the performance improvements become marginal, supporting our statement that only the most semantically related patches contribute significantly to the restoration.
(3) The randomly sampled strategy has a very stable and better performance compared to the fixed top-k manner especially when the inference $k$ is fixed to a small number (\emph{i.e.}, 64, 128, 256). 
We conclude that a random sampled strategy is more general and stable. It can also make the inference process more flexible regarding different computation resources. 
Meanwhile, we set a query region in the input and provided a detailed comparison from the attention-based activation map together with the input query region in Fig.~\ref{fig:dogvis_topk}. Fig.~\ref{fig:dogvis_topk}(a) shows the query region input. Fig.~\ref{fig:dogvis_topk}(b) displays the activation map generated using standard attention mechanisms. Fig.~\ref{fig:dogvis_topk}(c-f) illustrate activation maps using our key-semantic dictionary with different $k$ values ([8, 16, 64, 256]) during inference. The comparisons indicate that increasing $k$ allows for connections to more semantically related regions. However, when $k$ is set too high (\eg, $k$ = 256 as shown in Fig.~\ref{fig:dogvis_topk}(f)), the activation map may include some semantically unrelated regions. This aligns with the findings and the results depicted in Fig~\ref{fig:main_figure_K}, where increasing the $k$ beyond a certain point (e.g., from 396 to 512) does not further improve PSNR.
More ablation results can be found in our \textit{Appx.}~\ref{subsec:appx_moreabl} about the effect of the noise level and quality factor for denoising and JPEG CAR.

\noindent\textbf{Efficiency Analysis.} 
We compare our SemanIR method with four recent promising approaches on both 4x super-resolution (SR) and grayscale denoising tasks using the Urban100 dataset. Key metrics, including trainable parameters, FLOPs, runtime, and PSNR, are reported in Tab.~\ref{table:ab_efficiency}. The results show that HAT and SemanIR achieve top-tier PSNR performance, reaching 28.37 dB and 28.34 dB, respectively, while SemanIR-S is much faster and uses 41.7\% fewer parameters than HAT, making it more computationally efficient. SwinIR runs slightly faster than SemanIR-S but at the cost of a significant 0.89 dB loss in PSNR. In comparison, SemanIR-S offers better accuracy and speed than both CAT and HAT, solidifying its effectiveness. Furthermore, the comparison between SwinIR and SemanIR for denoising reveals that, although SwinIR has fewer trainable parameters, its FLOPs are substantially higher, indicating that SemanIR is more efficient in terms of computation. These results highlight that SemanIR-S strikes an optimal balance between performance and efficiency, making it highly competitive for both SR and denoising tasks.

\begin{table*}[t]
% \parbox{.48\linewidth}{
% \parbox{.38\linewidth}{
\centering
\caption{\textit{\textbf{Color image JPEG compression artifact removal}} results.} 
\label{table:jpeg_compression_artifacts_removal_color}
\vspace{-2mm}
\setlength{\tabcolsep}{2pt}
\setlength{\extrarowheight}{0.9pt}
\scalebox{0.69}{
\begin{tabular}{c | c | c c | c c | c c | c c | c c || c c | c c | c c}
\toprule[0.15em]
\multirow{2}{*}{Set} & \multirow{2}{*}{QF} & \multicolumn{2}{c|}{JPEG}  & \multicolumn{2}{c|}{\textcolor{magenta}{\textdagger}\makecell{QGAC \\ \cite{ehrlich2020quantization}}} & \multicolumn{2}{c|}{\textcolor{magenta}{\textdagger}\makecell{FBCNN \\ \cite{jiang2021towards}}} & \multicolumn{2}{c|}{\textcolor{magenta}{\textdagger}\makecell{DRUNet \\ \cite{zhang2021plug}}} & \multicolumn{2}{c||}{\textcolor{magenta}{\textdagger}\makecell{SemanIR\\ (Ours)}} & \multicolumn{2}{c|}{\makecell{SwinIR \\ \cite{liang2021swinir}}} & \multicolumn{2}{c|}{\makecell{GRL-S \\ \cite{li2023efficient}}} & \multicolumn{2}{c}{\makecell{SemanIR \\ (Ours)}} \\ \cline{3-18}
% & & PSNR/SSIM & PSNR/SSIM & PSNR/SSIM & PSNR/SSIM \\
& & PSNR$\uparrow$ & SSIM$\uparrow$ & PSNR$\uparrow$ & SSIM$\uparrow$ & PSNR$\uparrow$ & SSIM$\uparrow$ & PSNR$\uparrow$ & SSIM$\uparrow$ & PSNR$\uparrow$ & SSIM$\uparrow$ & PSNR$\uparrow$ & SSIM$\uparrow$ & PSNR$\uparrow$ & SSIM$\uparrow$ & PSNR$\uparrow$ & SSIM$\uparrow$  \\
\hline
{\multirow{4}{*}{\rotatebox[origin=c]{90}{{LIVE1}}}}
	&10	&25.69	&0.7430		&27.62	&0.8040		&\sotab{27.77}	&0.8030		&27.47	&\sotab{0.8045}		&\sotaa{28.19}	&\sotaa{0.8146}		&28.06	&0.8129		&\sotab{28.13}	&\sotab{0.8139}		&\sotaa{28.31}	&\sotaa{0.8176}		\\
	&20	&28.06	&0.8260		&29.88	&0.8680		&30.11	&0.8680		&\sotab{30.29}	&\sotab{0.8743}		&\sotaa{30.53}	&\sotaa{0.8781}		&30.44	&0.8768		&\sotab{30.49}	&\sotab{0.8776}		&\sotaa{30.61}	&\sotaa{0.8792}		\\
	&30	&29.37	&0.8610		&31.17	&0.8960		&31.43	&0.8970		&\sotab{31.64}	&\sotab{0.9020}		&\sotaa{31.89}	&\sotaa{0.9051}		&31.81	&0.9040		&\sotab{31.85}	&\sotab{0.9045}		&\sotaa{31.94}	&\sotaa{0.9058}		\\
	&40	&30.28	&0.8820		&32.05	&0.9120		&32.34	&0.9130		&\sotab{32.56}	&\sotab{0.9174}		&\sotaa{32.81}	&\sotaa{0.9201}		&32.75	&0.9193		&\sotab{32.79}	&\sotab{0.9195}		&\sotaa{32.85}	&\sotaa{0.9204}		\\ \hline
{\multirow{4}{*}{\rotatebox[origin=c]{90}{{BSDS500}}}}
    &10	&25.84	&0.7410		&27.74	&\sotab{0.8020}		&\sotab{27.85}	&0.7990		&27.62	&0.8001		&\sotaa{28.25}	&\sotaa{0.8076}		&28.22	&0.8075		&\sotab{28.26}	&\sotab{0.8083}		&\sotaa{28.37}	&\sotaa{0.8102}		\\
	&20	&28.21	&0.8270		&30.01	&0.8690		&30.14	&0.8670		&\sotab{30.39}	&\sotab{0.8711}		&\sotaa{30.55}	&\sotaa{0.8738}		&30.54	&0.8739		&\sotab{30.57}	&\sotab{0.8746}		&\sotaa{30.63}	&\sotaa{0.8750}		\\
	&30	&29.57	&0.8650		&31.33	&0.8980		&31.45	&0.8970		&\sotab{31.73}	&\sotab{0.9003}		&\sotaa{31.90}	&\sotaa{0.9026}		&31.90	&0.9025		&\sotab{31.92}	&\sotab{0.9030}		&\sotaa{31.96}	&\sotaa{0.9035}		\\
	&40	&30.52	&0.8870		&32.25	&0.9150		&32.36	&0.9130		&\sotab{32.66}	&\sotab{0.9168}		&\sotaa{32.84}	&\sotaa{0.9190}		&32.84	&0.9189		&\sotab{32.86}	&\sotab{0.9192}		&\sotaa{32.88}	&\sotaa{0.9193}		\\

\bottomrule[0.15em]
\end{tabular}}
% }
\vspace{-1mm}
\end{table*}
\begin{figure}[!t]
    \centering
    \includegraphics[width=1.0\linewidth]{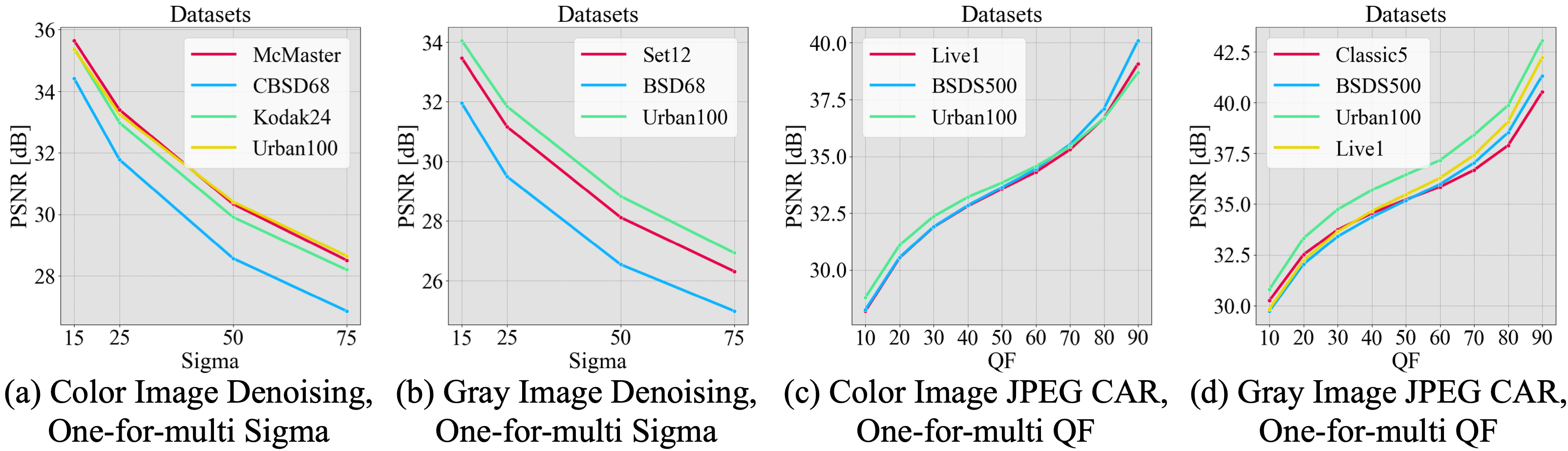}
    \vspace{-4mm}
    \caption{One model is trained to handle multi-degradations for denoising (a-b) and JPEG CAR (c-d).}
    \label{fig:one4m}
    \vspace{-4mm}
\end{figure}
\begin{table*}[!t]
\parbox{.50\linewidth}{
\scriptsize\begin{center}
\caption{\textit{\textbf{Gray image denoising}} PSNR.}
\label{table:ab_graydn_window}
\vspace{-1mm}
\setlength{\extrarowheight}{2pt}
\setlength{\tabcolsep}{5pt}
\scalebox{0.9}{
\begin{tabular}{l | c c c}
\toprule[0.15em]
Window Size	&Set12	& BSD68 & Urban100	\\ \hline
8   &31.01	&29.49	& 31.33 \\
16  &31.06	&29.51	& 31.45 \\
32	&31.17	&29.50	& 31.88 \\							\bottomrule[0.15em]
\end{tabular}
}
\end{center}
}
\parbox{.5\linewidth}{
\begin{center}
\caption{\textit{\textbf{Color image denoising}} PSNR.}
\label{table:ab_colordn_window}
\vspace{-1mm}
\setlength{\extrarowheight}{2pt}
\setlength{\tabcolsep}{5pt}
\scalebox{0.7}{
\begin{tabular}{l | c c c c}
\toprule[0.15em]
Window Size	& Mcmaster & CBSD68 & Kodak24 & Urban100	\\ \hline
8   & 33.20	& 31.72	& 32.84	& 32.89 \\
16  & 33.24	& 31.74 & 32.86	& 32.95	\\
32	& 33.38	& 31.75	& 32.97 & 33.27	\\							\bottomrule[0.15em]
\end{tabular}
}
\end{center}
}
\vspace{-7mm}
\end{table*}
\noindent\textbf{The Impact of One Model is Trained to Handle Multiple Degradation Levels.} 
The experiments were conducted for both denoising and JPEG CAR with both color and grayscale images. 
For denoising, $\sigma$ is set to $[15, 25, 50, 75]$. For JEPG CAR, QF is set to $[10, 20, 30, 40, 50, 60, 70, 80, 90]$. 
The results in Fig.~\ref{fig:one4m} indicate that 
the PSNR for both tasks across all the datasets, under both color and grayscale settings, decreases when the degraded level increases. It means that maintaining a decent generalization ability with one model to handle multiple degradation levels is not easy. Meanwhile, the proposed SemanIR can still outperform other methods on various tasks (See Tab.~\ref{table:denoising} and Tab.~\ref{table:jpeg_compression_artifacts_removal_color}), which means that capturing the key semantics is also essential for model's generalization ability.

\noindent\textbf{The Impact of the window size.} 
The windows indeed contain mixed information from different semantic parts. Yet, it is precisely this semantic distinction that motivates us to develop a selection mechanism for semantic information using KNN. 
We have conducted ablation studies of the window size (\ie, on both gray and color image denoising with $\sigma=25$). The results are summarized in Tab.~\ref{table:ab_graydn_window} and Tab.~\ref{table:ab_colordn_window}. 
With the increase of the window size, the semantic relevant information for each token is increased, thus leading to a PSNR gain for different IR tasks. 
For gray image denoising, larger window sizes lead to improved PSNR performance, consistently yielding higher PSNR values across different datasets. 
Similarly, in color image denoising, larger window sizes also result in better outcomes. 
For example, in the McMaster dataset, the PSNR increases from 33.20 dB with a window size of 8 to 33.38 dB with a window size of 32. 
These results suggest that larger window sizes enhance performance by capturing more contextual information.

\subsection{Evaluation of SemanIR on Various IR Tasks}
\noindent\textbf{Evaluation on Image deblurring.} Tab.~\ref{table:motion_deblurring} shows the quantitative results for single image motion deblurring on synthetic datasets ({GoPro} \cite{nah2017deep}, {HIDE} \cite{shen2019human}). 
Compared to the previous state-of-the-art Restormer~\cite{zamir2022restormer}, our SemanIR achieves significant PSNR improvement (\ie, 0.52 dB) on the GoPro dataset and the second-best on the HIDE dataset. The visual results are shown in the \textit{Appx.}~\ref{sec:appx_results}.

\begin{table*}[t]
% \parbox{.435\linewidth}{
\centering
\caption{\textit{\textbf{Color and grayscale image denoising}} PSNR results. 
% Both model complexity and accuracy are shown for better comparison.
}
\label{table:denoising}
\vspace{-1mm}
\setlength{\extrarowheight}{0.9pt}
\setlength{\tabcolsep}{2pt}
\scalebox{0.65}{
\begin{tabular}{l | r | c c c | c c c | c c c || c c c | c c c | c c c }
\toprule[0.15em]
\multirow{3}{*}{\textbf{Method}} & \multirow{3}{*}{\# \textbf{P}} & \multicolumn{9}{c||}{\textbf{Color}} & \multicolumn{9}{c}{\textbf{Grayscale}} \\ \cline{3-20}
& & \multicolumn{3}{c|}{\textbf{CBSD68}} & \multicolumn{3}{c|}{\textbf{McMaster}} & \multicolumn{3}{c||}{\textbf{Urban100}}  & \multicolumn{3}{c|}{\textbf{Set12}} & \multicolumn{3}{c|}{\textbf{BSD68}} & \multicolumn{3}{c}{\textbf{Urban100}} \\
 % \cline{3-14} \cline{17-25}
        &  & $\sigma$$=$$15$ & $\sigma$$=$$25$ & $\sigma$$=$$50$ & $\sigma$$=$$15$ & $\sigma$$=$$25$ & $\sigma$$=$$50$ & $\sigma$$=$$15$ & $\sigma$$=$$25$ & $\sigma$$=$$50$ & $\sigma$$=$$15$ & $\sigma$$=$$25$ & $\sigma$$=$$50$ & $\sigma$$=$$15$ & $\sigma$$=$$25$ & $\sigma$$=$$50$ & $\sigma$$=$$15$ & $\sigma$$=$$25$ & $\sigma$$=$$50$ \\ \hline
\textcolor{magenta}{\textdagger}DnCNN~\cite{kiku2016beyond}	&0.56	&33.90	&31.24	&27.95	&33.45	&31.52	&28.62	&32.98	&30.81	&27.59	&32.67	&30.35	&27.18	&31.62	&29.16	&26.23	&32.28	&29.80	&26.35	\\				
\textcolor{magenta}{\textdagger}FFDNet~\cite{zhang2018ffdnet}	&0.49	&33.87	&31.21	&27.96	&34.66	&32.35	&29.18	&33.83	&31.40	&28.05	&32.75	&30.43	&27.32	&31.63	&29.19	&26.29	&32.40	&29.90	&26.50	\\				
% \textcolor{magenta}{\textdagger}IRCNN	&0.19	&33.86	&31.16	&27.86	&34.58	&32.18	&28.91	&33.78	&31.20	&27.70	&32.76	&30.37	&27.12	&31.63	&29.15	&26.19	&32.46	&29.80	&26.22	\\				
\textcolor{magenta}{\textdagger}DRUNet~\cite{zhang2021plug}	&32.64	&34.30	&\sotab{31.69}	&28.51	&35.40	&33.14	&30.08	&34.81	&32.60	&29.61	&33.25	&30.94	&27.90	&\sotab{31.91}	&29.48	&26.59	&33.44	&31.11	&27.96	\\				
\textcolor{magenta}{\textdagger}Restormer~\cite{zamir2022restormer}	&26.13	&\sotab{34.39}	&\sotaa{31.78}	&\sotaa{28.59}	&\sotab{35.55}	&\sotab{33.31}	&\sotab{30.29}	&\sotab{35.06}	&\sotab{32.91}	&\sotab{30.02}	&\sotab{33.35}	&\sotab{31.04}	&\sotab{28.01}	&\sotaa{31.95}	&\sotaa{29.51}	&\sotaa{26.62}	&\sotab{33.67}	&\sotab{31.39}	&\sotab{28.33}	\\				
\textcolor{magenta}{\textdagger}SemanIR (Ours)	&25.82	&\sotaa{34.42}	&\sotaa{31.78}	&\sotab{28.57}	&\sotaa{35.65}	&\sotaa{33.40}	&\sotaa{30.34}	&\sotaa{35.37}	&\sotaa{33.26}	&\sotaa{30.41}	&\sotaa{33.47}	&\sotaa{31.16}	&\sotaa{28.12}	&\sotaa{31.95}	&\sotab{29.49}	&\sotab{26.54}	&\sotaa{34.05}	&\sotaa{31.84}	&\sotaa{28.83}	\\				
\hline																				
DnCNN~\cite{kiku2016beyond}	&0.56	&33.90	&31.24	&27.95	&33.45	&31.52	&28.62	&32.98	&30.81	&27.59	&32.86	&30.44	&27.18	&31.73	&29.23	&26.23	&32.64	&29.95	&26.26	\\				
% RNAN~\cite{zhang2019residual}	&8.96	&-	&-	&28.27	&-	&-	&29.72	&-	&-	&29.08	&-	&-	&27.70	&-	&-	&26.48	&-	&-	&27.65	\\				
% IPT~\cite{chen2021pre}	&115.33	&-	&-	&28.39	&-	&-	&29.98	&-	&-	&29.71	&-	&-	&-	&-	&-	&-	&-	&-	&-	\\				
EDT-B~\cite{li2021efficient}	&11.48	&34.39	&31.76	&28.56	&35.61	&33.34	&30.25	&35.22	&33.07	&30.16	&-	&-	&-	&-	&-	&-	&-	&-	&-	\\

DRUNet~\cite{zhang2021plug}	&32.64	&34.30	&31.69	&28.51	&35.40	&33.14	&30.08	&34.81	&32.60	&29.61	&33.25	&30.94	&27.90	&31.91	&29.48	&26.59	&33.44	&31.11	&27.96	\\				
SwinIR~\cite{liang2021swinir}	&11.75	&\sotab{34.42}	&31.78	&28.56	&35.61	&33.20	&30.22	&35.13	&32.90	&29.82	&33.36	&31.01	&27.91	&\sotab{31.97}	&29.50	&26.58	&33.70	&31.30	&27.98	\\	
Restormer~\cite{zamir2022restormer}	&26.13	&34.40	&31.79	&\sotab{28.60}	&35.61	&33.34	&30.30	&35.13	&32.96	&30.02	&33.42	&31.08	&28.00	&31.96	&\sotab{29.52}	&26.62	&33.79	&31.46	&28.29	\\				
			
Xformer~\cite{zhang2023xformer}	& 25.23	&\sotaa{34.43}	&\sotaa{31.82}	&\sotaa{28.63}	&\sotaa{35.68}	&\sotaa{33.44}	&\sotab{30.38}	&\sotab{35.29}	&\sotab{33.21}	&\sotab{30.36}	&\sotab{33.46}	&\sotab{31.16}	&\sotab{28.10}	&\sotaa{31.98}	&\sotaa{29.55}	&\sotaa{26.65}	&\sotab{33.98}	&\sotab{31.78}	&\sotab{28.71}	\\								
SemanIR (Ours)	& 25.82	&\sotaa{34.43}	&\sotab{31.79}	&\sotab{28.60}	&\sotab{35.65}	&\sotab{33.43}	&\sotaa{30.38}	&\sotaa{35.38}	&\sotaa{33.29}	&\sotaa{30.51}	&\sotaa{33.48}	&\sotaa{31.18}	&\sotaa{28.14}	&\sotab{31.97}	&\sotab{29.52}	&\sotab{26.53}	&\sotaa{34.09}	&\sotaa{31.87}	&\sotaa{28.86}	\\				
\bottomrule[0.15em]
\end{tabular}}
% }
\vspace{-4mm}
\end{table*}
\begin{table*}[!t]
\parbox{.55\linewidth}{
\scriptsize\begin{center}
\caption{\textit{\textbf{IR in adverse weather conditions}}.}
\label{table:weather}
\vspace{0mm}
\setlength{\extrarowheight}{1.8pt}
\setlength{\tabcolsep}{1pt}
\scalebox{0.9}{
\begin{tabular}{c|cc|cc|cc}
\toprule[0.15em]
    \multirow{2}{*}{\textbf{Type}} & \multicolumn{2}{c|}{\textbf{Test1 (rain+fog)}} & \multicolumn{2}{c|}{\textbf{SnowTest100k-L}} & \multicolumn{2}{c}{\textbf{RainDrop}}\\\cline{2-7}
    & Method     & PSNR$\uparrow$        & Method  & PSNR$\uparrow$          & Method  & PSNR$\uparrow$             \\ \hline
    \multirow{4}{*}{\rotatebox[origin=c]{90}{\makecell{Task \\ Specific}}}
    & pix2pix~\cite{isola2017image}        & 19.09       & DesnowNet~\cite{liu2018desnownet}    & 27.17            & AttGAN~\cite{qian2018attentive}    & 30.55                 \\
    & HRGAN~\cite{li2019heavy}        & 21.56       & JSTASR~\cite{chen2020jstasr}      & 25.32            & Quan~\cite{quan2019deep}    & 31.44                  \\
    & SwinIR~\cite{liang2021swinir}       & 23.23       & SwinIR     & 28.18            & SwinIR  & 30.82                \\
    & MPRNet~\cite{zamir2021multi}        & 21.90       & DDMSNET~\cite{zhang2021deep}     & 28.85            & CCN~\cite{quan2021removing}  & 31.34                \\ \hline 
    \multirow{3}{*}{\rotatebox[origin=c]{90}{\makecell{Multi\\ Task}}}     
    & All-in-One~\cite{li2020all}     & 24.71       & All-in-One  & 28.33            & All-in-One  & \sotaa{31.12}                \\
    & TransWea.~\cite{valanarasu2022transweather}  & \sotab{27.96}       & TransWea.& \sotab{28.48}            & TransWea.       & 28.84      \\ 
    & SemanIR (Ours)    & \sotaa{29.57}       & SemanIR (Ours)& \sotaa{30.76}              & SemanIR (Ours)       & \sotab{30.82}   \\
    \bottomrule[0.15em]
    
\end{tabular}
}
\end{center}
}
\parbox{.45\linewidth}{
\begin{center}
\scriptsize\caption{\textit{\textbf{Image demosaicking}} PSNR results.}
\label{table:demosaicking}
\vspace{0mm}
\setlength{\extrarowheight}{0.83pt}
\setlength{\tabcolsep}{5pt}
\scalebox{0.9}{
\begin{tabular}{l | c c}
\toprule[0.15em]
Datasets	&Kodak	&McMaster	\\ \hline
Matlab	&35.78	&34.43	\\
MMNet~\cite{kokkinos2019iterative}	&40.19	&37.09	\\
DDR~\cite{wu2016demosaicing} &41.11	&37.12	\\
DeepJoint~\cite{gharbi2016deep}	&42.00	&39.14	\\
RLDD~\cite{guo2020residual}	&42.49	&39.25	\\
DRUNet~\cite{zhang2021plug}	&42.68	&39.39	\\
RNAN~\cite{zhang2019residual}	&43.16	&39.70	\\
GRL~\cite{li2023efficient}	&\sotab{43.57}	&\sotab{40.22}	\\
SemanIR (Ours)	&\sotaa{43.62}	&\sotaa{40.68}	\\																			\bottomrule[0.15em]
\end{tabular}
}
\end{center}
}
\vspace{-6mm}
\end{table*}

\noindent\textbf{Evaluation on JPEG CAR.} 
The experiments for color images are conducted with 4 image quality factors ranging from 10 to 40 under two settings (\ie, \textcolor{magenta}{\textdagger} a single model is trained to handle multiple quality factors, and each model for each quality). 
The quantitative results shown in Tab.~\ref{table:jpeg_compression_artifacts_removal_color} indicate that our SemanIR achieves the best results on all the test sets across various quality factors among all the comparison methods for the color images. The visual comparisons in the \textit{Appx.}~\ref{sec:appx_results} further supports the effectiveness of our method.

\noindent\textbf{Evaluation on Image Denoising.} We show color and grayscale image denoising results in Tab.~\ref{table:denoising} under two settings (\emph{i.e.}, \textcolor{magenta}{\textdagger} one model for all noise levels $\sigma = \{15, 25, 50\}$ and each model for each noise level). 
For a fair comparison, both parameters and accuracy are reported for all the methods.
For \textcolor{magenta}{\textdagger}, our SemanIR performs better on all test sets for color and grayscale image denoising than others. 
It is worth noting that we outperform DRUNet and Restormer with lower trainable parameters. 
For another setting, the proposed SemanIR also archives better results on CBSD68 and Urban100 for color image denoising, and on Set12 and Urban100 for grayscale denoising. 
These interesting comparisons validate the effectiveness of the proposed SemanIR and also indicate that our method has a higher generalization ability. 
The visual results in \textit{Appx.}~\ref{sec:appx_results} also support that the proposed SemanIR can remove heavy noise corruption and preserve high-frequency image details, resulting in sharper edges and more natural textures without over-smoothness or over-sharpness problems.

\begin{table*}[t]
\scriptsize
\setlength{\tabcolsep}{4pt}
\setlength{\extrarowheight}{0.9pt}
\setlength{\abovecaptionskip}{0.1cm}
\begin{center}
\caption{\textbf{\textit{Classical image SR}} results. Both lightweight and accurate models are summarized.}
\vspace{1mm}
\label{tab:sr_results}
\scalebox{0.96}{
\begin{tabular}{l|c|r|cc|cc|cc|cc|cc}
\toprule[0.15em]
\multirow{2}{*}{\textbf{Method}} & \multirow{2}{*}{\textbf{Scale}} & {\textbf{Params}} &  \multicolumn{2}{c|}{\textbf{Set5}} &  \multicolumn{2}{c|}{\textbf{Set14}} &  \multicolumn{2}{c|}{\textbf{BSD100}} &  \multicolumn{2}{c|}{\textbf{Urban100}} &  \multicolumn{2}{c}{\textbf{Manga109}} 
\\
\cline{4-13}
&  & \multicolumn{1}{c|}{\textbf{[M]}} & PSNR$\uparrow$ & SSIM$\uparrow$ & PSNR$\uparrow$ & SSIM$\uparrow$ & PSNR$\uparrow$ & SSIM$\uparrow$ & PSNR$\uparrow$ & SSIM$\uparrow$ & PSNR$\uparrow$ & SSIM$\uparrow$
\\
\hline
RCAN~\cite{zhang2018rcan}&	2$\times$&	15.44&	38.27&	0.9614&	34.12&	0.9216&	32.41&	0.9027&	33.34&	0.9384&	39.44&	0.9786\\
SAN~\cite{dai2019SAN}&	2$\times$&	15.71&	38.31&	0.9620&	34.07&	0.9213&	32.42&	0.9028&	33.10&	0.9370&	39.32&	0.9792\\
HAN~\cite{niu2020HAN}&	2$\times$&	63.61&	38.27&	0.9614&	34.16&	0.9217&	32.41&	0.9027&	33.35&	0.9385&	39.46&	0.9785\\
% NLSA~\cite{mei2021NLSA}&	2$\times$&	42.63&	38.34&	0.9618&	34.08&	0.9231&	32.43&	0.9027&	33.42&	0.9394&	39.59&	0.9789\\
% DBPN~\cite{haris2018DBPN}&	2$\times$&	1.26&	38.09&	0.9600&	33.85&	0.9190&	32.27&	0.9000&	32.55&	0.9324&	38.89&	0.9775\\
IPT~\cite{chen2021pre}&	2$\times$&	115.48&	38.37&	-&	34.43&	-&	32.48&	-&	33.76&	-&	-&	-\\ \hline
% SwinIR~\cite{liang2021swinir}&	2$\times$&	 {0.88}&	38.14&	0.9611&	33.86&	0.9206&	32.31&	0.9012&	32.76&	0.9340&	39.12&	0.9783\\
SwinIR~\cite{liang2021swinir}&	2$\times$&	11.75&	38.42&	0.9623&	34.46&	0.9250&	32.53&	0.9041&	33.81&	0.9427&	39.92&	0.9797\\  
% EDT~\cite{li2021efficient}&	2$\times$&	0.92&	38.23&	0.9615&	33.99&	0.9209&	32.37&	0.9021&	32.98&	0.9362&	39.45&	0.9789\\
% GRL-T (ours) &	2$\times$&	 {0.89}&	38.27&	0.9627&	34.21&	0.9258&	32.42&	0.9056&	33.60&	0.9411&	39.61&	0.9790\\
% GRL-S (ours) &	2$\times$&	3.34&	38.37&	 {0.9632}&	34.64&	 {0.9280}&	32.52&	 {0.9069}&	 {34.36}&	 {0.9463}&	39.84&	0.9793\\
CAT-A~\citep{chen2022cross}    & 2$\times$ &16.46 & 38.51 & 0.9626 & 34.78 & 0.9265 & 32.59 & 0.9047 & 34.26 & 0.9440 & 40.10 & 0.9805 \\
ART~\cite{zhang2023accurate}	&2$\times$ &16.40	&38.56	&0.9629	&34.59	&0.9267	&32.58	&0.9048	&34.30	&0.9452	&40.24	&0.9808	\\		
EDT~\cite{li2021efficient}&	2$\times$&	11.48&	 \sotaa{38.63}&	 {0.9632}&	 {34.80}&	0.9273&	 {32.62}&	0.9052&	34.27&	\sotab{0.9456}&	 {40.37}&	 {0.9811}\\ 
% GRL-B~\cite{li2023efficient} &	2$\times$&	20.05&	 \sotaa{38.67}&	 \sotab{0.9647}&	 {35.08}&	 \sotab{0.9303}&	 {32.67}&	 \sotab{0.9087}&	 \sotaa{35.06}&	 \sotaa{0.9505}&	 \sotaa{40.67}&	 \sotab{0.9818}\\ 

SemanIR-S (Ours)	&2$\times$& 11.87	&38.57	&\sotab{0.9651}	&\sotab{34.99}	&\sotab{0.9300}	&\sotab{32.65}	&\sotab{0.9078}	&\sotab{34.86}	&\sotaa{0.9472}	&\sotab{40.45}	&\sotab{0.9824}			\\																			
SemanIR-B (Ours) &	2$\times$&	19.90&	 \sotab{38.61}&	 \sotaa{0.9654}&	 \sotaa{35.08}&	 \sotaa{0.9304}&	 \sotaa{32.69}&	 \sotaa{0.9084}&	 \sotaa{34.99}&	 
{0.9455}&	 \sotaa{40.59}&	 \sotaa{0.9830}\\ 
% Base Phase5 Results

\midrule[0.1em]

RCAN~\cite{zhang2018rcan}&	4$\times$&	15.59&	32.63&	0.9002&	28.87&	0.7889&	27.77&	0.7436&	26.82&	0.8087&	31.22&	0.9173\\
SAN~\cite{dai2019SAN}&	4$\times$&	15.86&	32.64&	0.9003&	28.92&	0.7888&	27.78&	0.7436&	26.79&	0.8068&	31.18&	0.9169\\
HAN~\cite{niu2020HAN}&	4$\times$&	64.20&	32.64&	0.9002&	28.90&	0.7890&	27.80&	0.7442&	26.85&	0.8094&	31.42&	0.9177\\
% NLSA~\cite{mei2021NLSA}&	4$\times$&	44.99&	32.59&	0.9000&	28.87&	0.7891&	27.78&	0.7444&	26.96&	0.8109&	31.27&	0.9184\\
% DBPN~\cite{haris2018DBPN}&	4$\times$&	2.21&	32.47&	0.8980&	28.82&	0.7860&	27.72&	0.7400&	26.38&	0.7946&	30.91&	0.9137\\
IPT~\cite{chen2021pre}&	4$\times$&	115.63&	32.64&	-&	29.01&	-&	27.82&	-&	27.26&	-&	-&	-\\ \hline
% IPT~\cite{chen2021pre}&	$\times$3&	115.67&	34.81&	-&	30.85&	-&	29.38&	-&	29.49&	-&	-&	-\\ \hline
% RRDB~\cite{wang2018esrgan}&	4$\times$&	16.70&	32.73&	0.9011&	28.99&	0.7917&	27.85&	0.7455&	27.03&	0.8153&	31.66&	0.9196\\    \hline
% SwinIR~\cite{liang2021swinir}&	4$\times$&	 {0.90}&	32.44&	0.8976&	28.77&	0.7858&	27.69&	0.7406&	26.47&	0.7980&	30.92&	0.9151\\
SwinIR~\cite{liang2021swinir}&	4$\times$&	11.90&	32.92&	0.9044&	29.09&	0.7950&	27.92&	0.7489&	27.45&	0.8254&	32.03&	0.9260\\  

CAT-A~\citep{chen2022cross}      & 4$\times$ &16.60 & {33.08} & 0.9052 & 29.18 & 0.7960 & 27.99 & 0.7510 & 27.89 & 0.8339 & 32.39 & 0.9285 \\
ART~\cite{zhang2023accurate}	&4$\times$	& 16.55 &33.04	&0.9051	&29.16	&0.7958	&27.97	&0.751	&27.77	&0.8321	&32.31	&0.9283	\\		
EDT~\cite{li2021efficient}&	4$\times$&	11.63&	\sotab{33.06}&	0.9055&	 {29.23}&	0.7971&	 \sotaa{27.99}&	0.7510&	27.75&	\sotab{0.8317}&	 {32.39}&	 {0.9283}\\ 

SemanIR-S (Ours)	&4$\times$& 12.02	&33.02	&\sotab{0.9082}	&\sotab{29.29}	&\sotab{0.8026}	&27.96	&\sotab{0.7582}	&\sotab{28.34}	&\sotaa{0.8467}	&\sotab{32.48}	&\sotab{0.9322}			\\				
SemanIR-B (Ours) &	4$\times$&	20.04&	 \sotaa{33.08}&	 \sotaa{0.9090}&	 \sotaa{29.34}&	 \sotaa{0.8037}&	 \sotab{27.98}&	 \sotaa{0.7599}&	 \sotaa{28.51}&	 \sotaa{0.8467}&	 \sotaa{32.56}&	 \sotaa{0.9335}\\ 
\bottomrule[0.15em]
\end{tabular}
}
\end{center}
% \vspace{-2mm}
\end{table*}
\begin{figure*}[!t]
    \centering
    \includegraphics[width=0.99\linewidth]{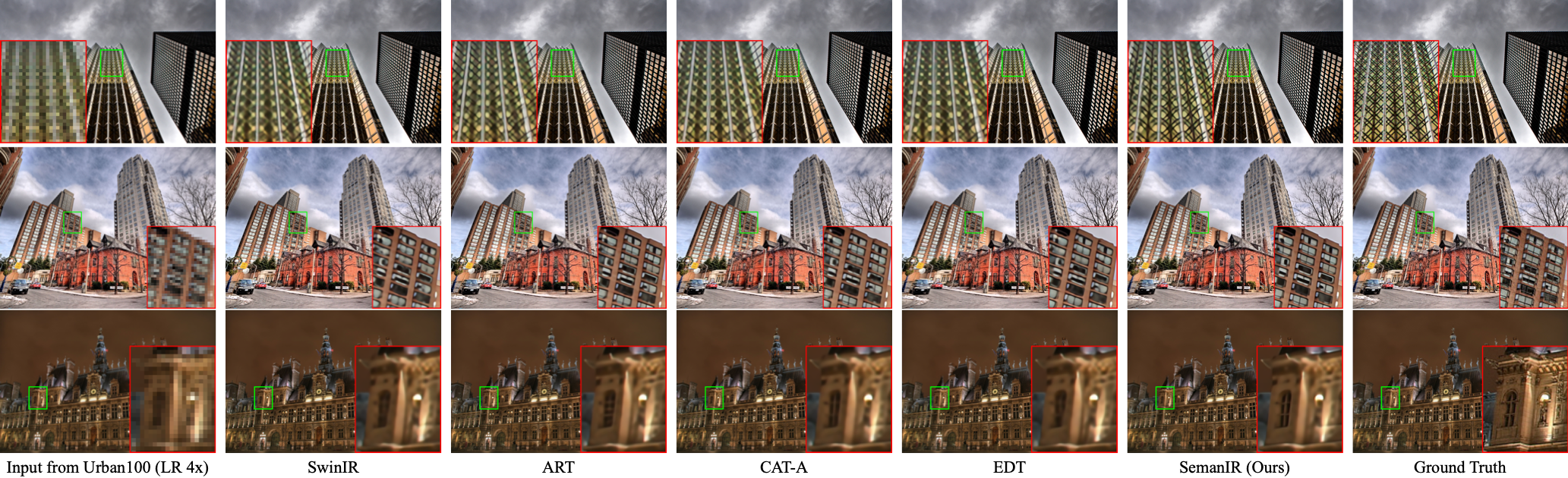}
    % \vspace{0mm}
    \caption{Visual comparison of classical image SR (4$\times$) on Urban100. Best viewed by zooming.}
    \label{fig:sr_x4}
    % \vspace{-3mm}
\end{figure*}

\noindent\textbf{Evaluation in AWC.}
We validate SemanIR in adverse weather conditions, including rain+fog (Test1), snow (SnowTest100K), and raindrops (RainDrop). 
PSNR is reported in Tab.~\ref{table:weather}. Our method achieves the best performance on Test1 (\emph{i.e.}, 5.76\% improvement) and SnowTest100k-L (\emph{i.e.} 8.01\% improvement), while the second-best PSNR on RainDrop compared to all other methods. 
See \textit{Appx.}~\ref{sec:appx_results} for 
Visual comparisons.

\noindent\textbf{Evaluation on Image Demosaicking.} 
The quantitative results shown in \ref{table:demosaicking} indicate that the proposed SemanIR performs best on both the Kodak and MaMaster test sets, especially 0.05dB and 0.45dB absolute improvement compared to the current state-of-the-art. 

\noindent\textbf{Evaluation on SR.} For the classical image SR, we compared our SemanIR with both recent lightweight and accurate SR models, and the quantitative results are shown in Tab.~\ref{tab:sr_results}. 
Compared to EDT, SemanIR-base achieves significant improvements on Urban100 (\ie, 0.72 dB and 0.76dB for 2$\times$ and 4$\times$ SR) and Manga109 datasets (\ie, 0.22dB and 0.17 dB for 2$\times$ and 4$\times$ SR). 
Even the SemanIR-small consistently ranks as the runner-up across the majority of test datasets, all while maintaining a reduced number of trainable parameters. 
Visual results in both Fig.~\ref{fig:sr_x4} and \textit{Appx.}~\ref{sec:appx_results} also validate the effectiveness of the proposed SemanIR. Specifically, it is clear from the zoomed part in Fig.~\ref{fig:sr_x4} that SemanIR can restore more details and structural content compared to other methods.

%-----------------------%
\section{Conclusion}
\label{sec:conclusion}
%-----------------------%
In this paper, we propose a novel approach, SemanIR, for ViTs-based image restoration, which experimentally validated that global cues are essential to restore degraded images well, but the most semantically related global cures play the 
major role.
Specifically, to capture the key semantics, we propose to construct a semantic dictionary (\ie, naively by self-similarity is enough) for storing only the most related $k$ semantic information and then use it as a reference for guiding the attention operation for making the attention operation pay more attention only to these key semantics. 
Furthermore, we share the key-semantic dictionary with all the upcoming transformer layers within the same stage since each stage of the transformer is typically at the same semantic level. This strategy significantly reduces the computational cost for IR and functions as loop optimization, continuously restoring degraded patches with their most semantically related patches, which share similar texture or structural information.
Extensive experiments on 6 IR tasks validated the effectiveness of SemanIR, demonstrating that our method achieves new state-of-the-art performance. 

\begin{ack}
We thank Danda Pani Paudel for the valuable discussions on this work. This work was partially supported by Shenzhen Innovation in Science and Technology Foundation for The Excellent Youth Scholars (No. {RCYX20231211090248064}), the National Natural Science Foundation of China (No. 62203476), the MUR PNRR project FAIR (PE00000013) funded by the NextGenerationEU, the PRIN project CREATIVE (Prot. 2020ZSL9F9), the EU Horizon project ELIAS (No. 101120237), and the Ministry of Education and Science of Bulgaria (support for INSAIT, part of the Bulgarian National Roadmap for Research Infrastructure).
\end{ack}

\bibliographystyle{plain}
\bibliography{egbib}

\appendix
\section{Experimental Protocals}
\label{sec:appx_experimental}

\subsection{Training/Testing Datasets}
\label{subsec:appx_dataset}
\noindent\textbf{JPEG compression artifact removal.}
For JPEG compression artifact removal, the JPEG image is compressed by the \texttt{cv2} JPEG compression function. The compression function is characterized by the quality factor. We investigated four compression quality factors including 10, 20, 30, and 40. The smaller the quality factor, the more the image is compressed, meaning a lower quality. 

\begin{itemize}
    \item  The training datasets: DIV2K~\cite{agustsson2017ntire}, Flickr2K~\cite{lim2017enhanced}, and WED~\cite{ma2016waterloo}. 

    \item  The test datasets: Classic5~\cite{foi2007Classic5}, LIVE1~\cite{sheikh2005live}, Urban100~\cite{huang2015single}, BSD500~\cite{arbelaez2010contour}. 
\end{itemize}

\noindent\textbf{Image Denoising.}
For image denoising, we conduct experiments on both color and grayscale image denoising. During training and testing, noisy images are generated by adding independent additive white Gaussian noise (AWGN) to the original images. The noise levels are set to $\sigma = 15, 25, 50$. We train individual networks at different noise levels. The network takes the noisy images as input and tries to predict noise-free images. 

\begin{itemize}
    \item  The training datasets: DIV2K~\cite{agustsson2017ntire}, Flickr2K~\cite{lim2017enhanced}, WED~\cite{ma2016waterloo}, and BSD400~\cite{martin2001database}. 

    \item  The test datasets for color image: CBSD68~\cite{martin2001database}, Kodak24~\cite{franzen1999kodak}, McMaster~\cite{zhang2011color}, and Urban100~\cite{huang2015single}.
    
    \item  The test datasets for grayscale image: Set12~\cite{zhang2017beyond}, BSD68~\cite{martin2001database}, and Urban100~\cite{huang2015single}.
\end{itemize}

\noindent\textbf{Image Demosaicking.}
For image demosaicking, the mosaic image is generated by applying a Bayer filter on the ground-truth image. Then the network try to restore high-quality image. The mosaic image is first processed by the default \texttt{Matlab} demosaic function and then passed to the network as input. 

\begin{itemize}
    \item The training datasets: DIV2K~\cite{agustsson2017ntire} and Flickr2K~\cite{lim2017enhanced}. 

    \item The test datasets: Kodak~\cite{franzen1999kodak}, McMaster~\cite{zhang2011color}. 
\end{itemize}

\noindent\textbf{IR in Adverse Weather Conditions.} 
For IR in adverse weather conditions, the model is trained on a combination of images degraded by a variety of adverse weather conditions. The same training and test dataset is used as in Transweather~\cite{valanarasu2022transweather}. The training data comprises 9,000 images sampled from Snow100K \cite{liu2018desnownet}, 1,069 images from Raindrop \cite{qian2018attentive}, and 9,000 images from Outdoor-Rain \cite{li2019heavy}. Snow100K includes synthetic images degraded by snow, Raindrop consists of real raindrop images, and Outdoor-Rain contains synthetic images degraded by both fog and rain streaks. The proposed method is tested on both synthetic and real-world datasets.

\begin{itemize}
    \item The comparison methods in Tab. 6 of our main manuscript: pix2pix~\cite{isola2017image}, HRGAN~\cite{li2019heavy},
    SwinIR~\cite{liang2021swinir}, All-in-One~\cite{li2020all}, Transweather~\cite{valanarasu2022transweather}, DesnowNet~\cite{liu2018desnownet}, JSTASR~\cite{chen2020jstasr}, DDMSNET~\cite{zhang2021deep}, Attn. GAN~\cite{qian2018attentive},~\cite{quan2019deep}, and CCGAN~\cite{quan2021removing}.

    \item The test datasets: test1 dataset~\cite{li2020all, li2019heavy}, the RainDrop test dataset~\cite{qian2018attentive}, and the Snow100k-L test. 
\end{itemize}

\noindent\textbf{Image SR.}
For image SR, the LR image is synthesized by \texttt{Matlab} bicubic downsampling function before the training. We investigated the upscalingg factors $\times2$, $\times3$, and $\times4$. 

\begin{itemize}
    \item  The training datasets: DIV2K~\cite{agustsson2017ntire} and Flickr2K~\cite{lim2017enhanced}. 

    \item  The test datasets: 
    Set5~\cite{bevilacqua2012low}, Set14~\cite{zeyde2010single}, BSD100~\cite{martin2001database}, Urban100~\cite{huang2015single}, and Manga109~\cite{matsui2017sketch}.
\end{itemize}

\noindent\textbf{Image Deblurring.}
For single-image motion deblurring, 

\begin{itemize}
    \item  The training datasets: GoPro~\cite{nah2017deep} . 

    \item  The test datasets: 
    GoPro~\cite{nah2017deep} and HIDE~\cite{shen2019human}.
\end{itemize}

\begin{figure*}[!t]
    \centering
    \vspace{0mm}
    \includegraphics[width=0.6\linewidth]{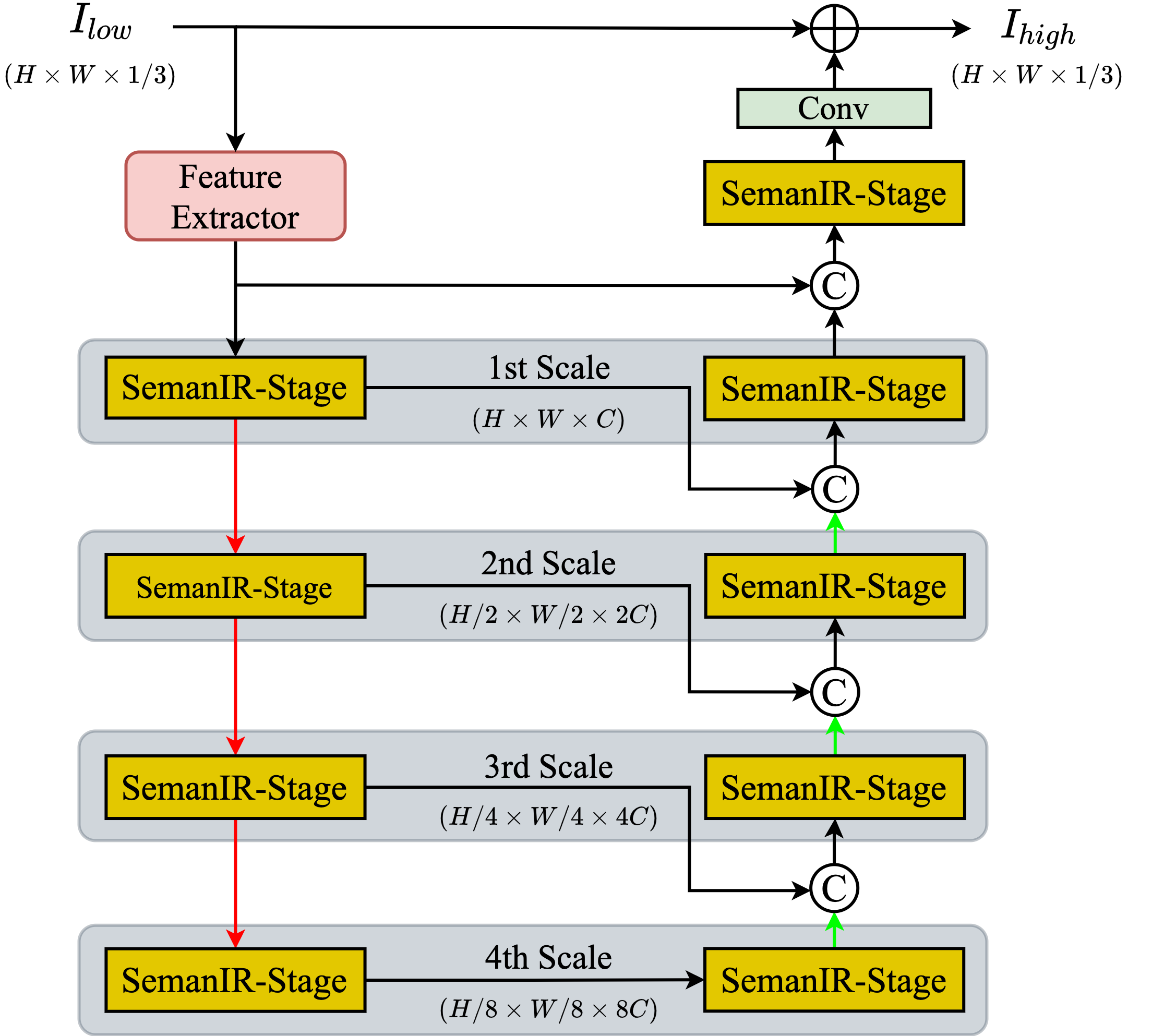}
    \caption{The U-shaped hierarchical architecture (Archi-V2) of the proposed SemanIR for Image Restoration. Note that this U-shaped one is used for image JPEG CAR, image denoising, image demosaicking, IR in AWC, and image deblurring. Symbols $\oplus$ and \textcircled{c} denote the element-wise addition and channel-wise concatenation. The downsample and upsample operations are denoted by \textcolor{red}{red} and \textcolor{green}{green} arrows.}
    \vspace{-3mm}
    \label{fig:ushape}
\end{figure*}

\begin{table*}[ht]
    \centering
    \caption{The details of the transformer stages and layers per stage of SemanIR for both architectures.}
    \setlength{\extrarowheight}{1pt}
    \setlength\tabcolsep{4pt} 
    \label{table:model_details}
    \scalebox{0.88}{
    \begin{tabular}{c|cc|ccc}
    \toprule[0.15em]
     & \multicolumn{2}{c|}{Archi-V1 (Columnar-shape)} & \multicolumn{3}{c}{Archi-V2 (U-shape)} \\ \hline
     & \multicolumn{1}{c|}{SemanIR-small} & SemanIR-base & \multicolumn{1}{c|}{Down Stages} & \multicolumn{1}{c|}{Up Stages} & Final Stage \\ \hline
    Num. of SemanIR Stages & \multicolumn{1}{c|}{6} & 8 & \multicolumn{1}{c|}{4} & \multicolumn{1}{c|}{4} & 1 \\
    Num. of SemanIR layer per stage & \multicolumn{1}{c|}{6} & 8 & \multicolumn{1}{c|}{6} & \multicolumn{1}{c|}{6} & 6 \\ \bottomrule[0.15em]
    \end{tabular}
    }
\end{table*}

\subsection{Model Architecture}
\label{subsec:appx_architecture}
In the proposed SemanIR, we adopt two kinds of base architecture \ie, the widely used multi-stage one shown in Fig.1 of our main manuscript (Archi-V1) and a U-shaped hierarchical one shown in Fig.~\ref{fig:ushape} (Archi-V2) for taking patterns of various scales into account (Note that 1/3 of $I_{low}$ and $I_{high}$ in Fig.~\ref{fig:ushape} denotes the grayscale/color image cases). This is consistent with previous methods such as Restormer~\cite{zamir2022restormer}, KiT~\cite{lee2022knn}, and NAFNet~\cite{chen2022simple}. 

Note that The feature extractor for both architectures is implemented as a simple convolution and converts the input image into feature maps. The image reconstructor for Archi-V1 takes the rich features calculated by
the previous operations and estimates a recovered image.

In addition to introducing the two base architectures of the proposed SemanIR, we have provided comprehensive details of its structure in Table~\ref{table:model_details}. This table outlines the number of SemanIR stages and the distribution of layers within each stage, offering a thorough understanding of our model's architecture.

\subsection{Efficiency Analysis}
\label{subsec:appx_efficiency}
\begin{table}[t]
\centering
\caption{The Computation Complexity Comparison.}
\label{table:efficiency}
\setlength{\extrarowheight}{1.5pt}
\setlength{\tabcolsep}{3pt}
\begin{tabular}{ccc}
\toprule[0.12em]
 & Time Complexity & Space Complexity \\ \hline
MSA & $\mathcal{O}(4HWC^{2} + 2(HW)^{2}C)$ & $\mathcal{O}(4HWC^{2} + 2h(HW)^{2}C)$ \\
W-MSA & $\mathcal{O}(4HWC^{2} + 2(M)^{2}HWC)$ & $\mathcal{O}(4HWC^{2} + 2h(M)^{2}HWC)$ \\
SemanIR MSA (Ours) & $\mathcal{O}(4HWC^{2} + 2kHWC)$ & $\mathcal{O}(4HWC^{2} + 2hkHWC)$ \\ 
\bottomrule[0.12em]
\end{tabular}
\end{table}

We provide a complexity comparison among the standard multi-head self-attention (MSA), the Window-wise MSA (W-MSA), and the proposed KeySemanIR MSA (SemanIR-MSA) in the Tab.~\ref{table:efficiency}. $(H, W, C)$ indicate the feature size, $M$ represents the window size, and $h$ denotes the number of heads. It is commonly demonstrated and proven that the complexity of the W-MSA is much lower than that of the standard MSA, \ie,

\begin{equation}
\begin{aligned}
\mathcal{O}(4HWC^{2} + 2(M)^{2}HWC) < \mathcal{O}(4HWC^{2} + 2(HW)^{2}C)
\end{aligned}
\end{equation}

To better understand the efficiency of the proposed method, it should be considered together with all transformer layers within a certain stage. Specifically, take one stage, which contains 6 transformer Layers, as an example (To simplify the illustration, we omit the convolution operation at the end of each stage). 

First, the total complexity of W-MSA within each stage can be calculated as:
\begin{equation}
\begin{aligned}
    \mathcal{O}(6 \times [4HWC^{2} + 2(M)^{2}HWC])
\end{aligned}
\end{equation}
Second, similarly, the complexity of the proposed SemanIR-MSA can be calculated as follows. ($\mathcal{O}(HWC)$ indicates the complexity of the key-semantic dictionary construction at the start of each transformer stage. All the other layers then share it, hence it is calculated only once).
\begin{equation}
\begin{aligned}
    \mathcal{O}(6 \times [4HWC^{2} + 2kHWC] + (HW)^{2}C)
\end{aligned}
\end{equation}

Third, a simple subtraction can be done as follows to validate that the proposed method is more efficient compared to the W-MSA within each Transformer stage:
\begin{equation}
\begin{aligned}
    \mathcal{O}(6 \times [4HWC^{2} + 2(M)^{2}HWC] - (6 \times [4HWC^{2} + 2kHWC] + (HW)^{2}C)) \\ = \mathcal{O}((12M^{2} - 12k - HW)HWC)
\end{aligned}
\end{equation}
In the last equation, to provide further clarity, let’s consider a common setting where the window size $M = 7$ and the patch size is 16. The height $H$ and the width $W$ of the feature map are 64. The number of pixels within the window is approximately (7 $\times$ 7) $\times$ (16 $\times$ 16). In the proposed SemanIR, the k value is set to 512 or randomly sampled from [64, 128, 256, 384, 512]. To this end, we have:
\begin{equation}
\begin{aligned}
  \mathcal{O}((12M^{2} - 12k - HW)HWC) &= \mathcal{O}(12 \times (7 \times 7) \times(16 \times 16) - 12 \times 512 - 64 \times 64) \\&= \mathcal{O}(150528 - 6144 - 4096) >> 0
\end{aligned}
\end{equation}
This shows that the complexity is significantly greater than zero.

Based on the above analysis, it can be concluded that, together with the proposed transformer layer, constructing the key-semantic dictionary at the start of each stage leads to greater efficiency.

\subsection{Training Details}
\label{subsec:appx_training}
Our method explores \textbf{6} IR tasks, and the training settings vary slightly for each task. These differences encompass the architecture of the proposed SemanIR, variations in the choice of the optimizer, and loss functions. 
Each experiments are conducted on 4 NVIDIA Tesla V100 32G GPUs.

\begin{figure*}[!t]
    \centering
    \includegraphics[width=1.0\linewidth]{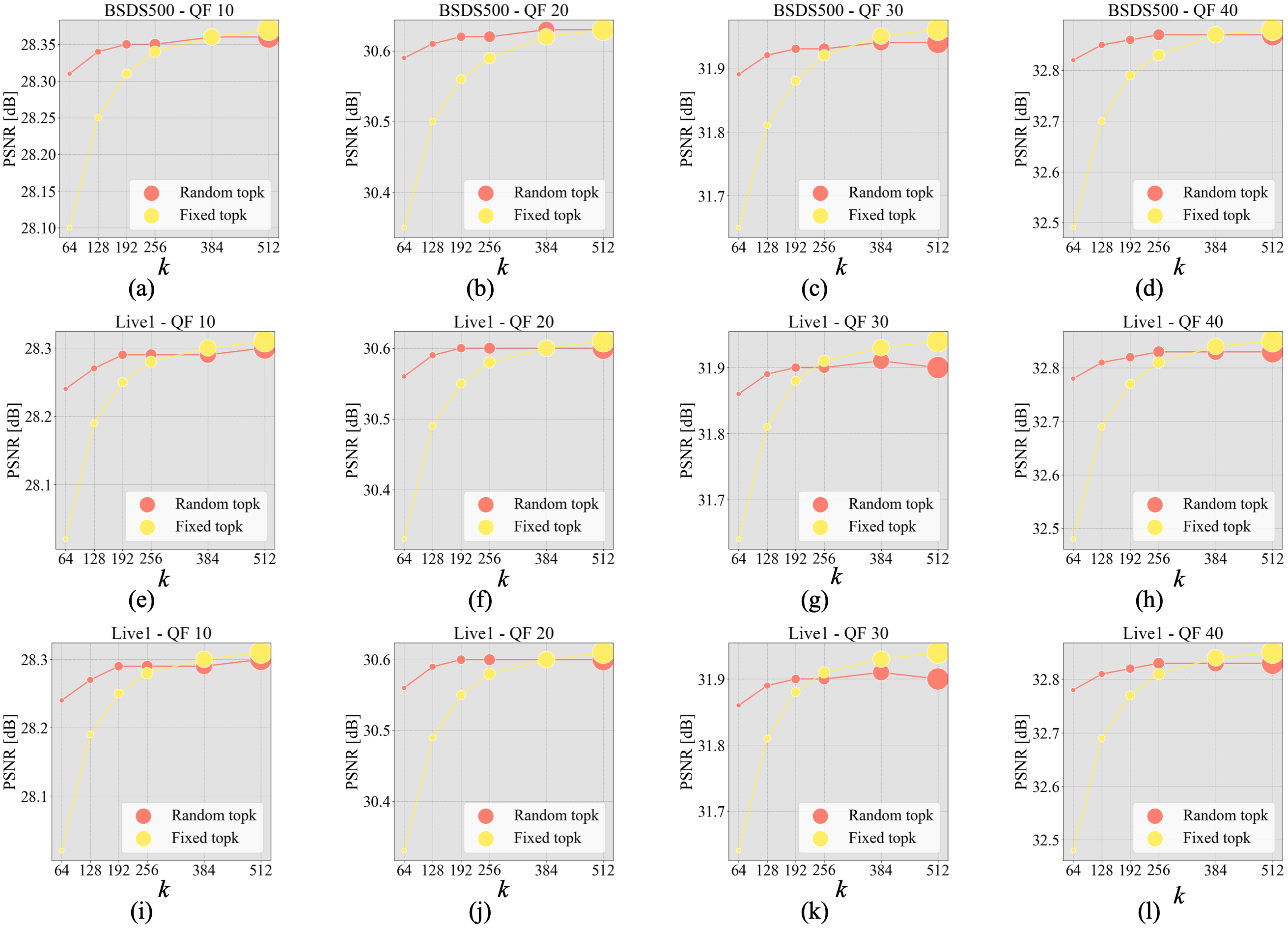}
    \vspace{-3mm}
    \caption{Ablation study on the impact of $k$ for \textbf{Color JPEG CAR} on CBSD68 (a-d), Live (e-h), and Urban100 (i-l) datasets with $QF$ = \{10, 20, 30, 40\}.}
    \label{fig:supple_fig_K_jpeg_color}
    \vspace{-2mm}
\end{figure*}

\noindent \textbf{Architecture.} We use the columnar multi-stage architecture (without changing the feature map resolution and number of channels) for image SR and the U-shaped architecture for the other tasks including image denoising, image deblurring, and other tasks. The strategy of using multiple architectures is also explored by the previous method~\cite{chen2023activating, chen2022simple}. 

\noindent \textbf{Optimizer.} We adopt the same optimizer as all other comparison methods, \ie, Adam~\cite{kingma2014adam}, for IR in AWC, and AdamW~\cite{loshchilov2018decoupled} for the rest IR tasks.

\noindent \textbf{Loss Function.} We adopt the same loss function as all other comparison methods, \ie, smooth L1 loss and VGG loss~\cite{johnson2016perceptual,simonyan2015very} for IR in AWC, the Charbonnier loss for Deblurring, and L1 loss for the rest IR tasks.

\noindent \textbf{Batch Size and Patch Size.} 
We keep the similar batch size as other comparison methods, \ie, (Batch size = 16, Patch Size = 64) for JPEG CAR, denoising, demosaicking, and SR. (Batch Size = 32, Patch Size = 16) for IR in AWC. (Batch Size = 8, Patch Size = 192) for deblurring.

\noindent \textbf{Learning Rate Schedule.} For all the IR tasks, similar to other comparison methods, we set the initial learning rate to 2 $\times$ 10$^{-4}$, and then the half-decay is adopted during the training. Note that the training iteration for JPEG CAR, denoising, demosaicking, and SR is set to 1M. For IR in AWC and debluriing, it is set to 750K.

\subsection{Evaluation Introduction}
\label{subsec:appx_eval}
Note that the results of all the comparison methods are reported from their original papers. The details of the evaluation metric (\ie, SSIM, PSNR) are described as follows:

\noindent\textbf{JPEG compression artifact removal.} 
For color image JPEG compression artifact removal, the PSNR is reported on the RGB channels while for grayscale image JPEG compression artifact removal, the PSNR is reported on the Y channel.

\noindent\textbf{Image Denoising.}
For color image denoising, the PSNR is reported on the RGB channels while for grayscale image denoising, the PSNR is reported on the Y channel.

\noindent\textbf{Image Demosaicking.}
For the comparison between different methods, PSNR is reported on the RGB channels.

\noindent\textbf{IR in Adverse Weather Conditions.} We adopted the same PSNR evaluation metric used in Transweather~\cite{valanarasu2022transweather}.

\noindent\textbf{Image SR.}
The PSNR is reported on the Y channel.

\noindent\textbf{Image Deblurring.}
The PSNR and SSIM on the RGB channels are reported.

\begin{figure*}[!t]
    \centering
    \includegraphics[width=1.0\linewidth]{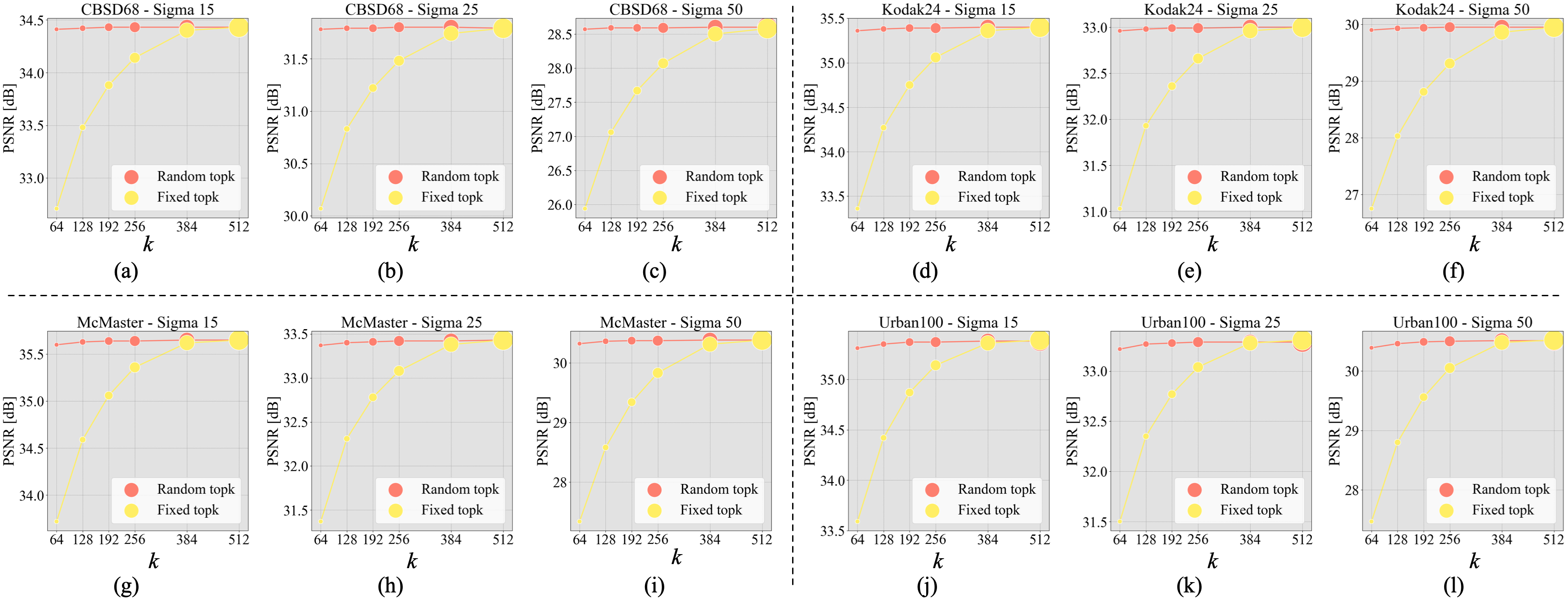}
    \vspace{-3mm}
    \caption{Ablation study on the impact of $k$ for \textbf{Color Image Denoising} on CBSD68 (a-c), Kodak24 (d-f), McMaster (g-i), and Urban100 (j-l) datasets with $\sigma$ = \{15, 25, 50\}.}
    \label{fig:supple_fig_K_dn_color}
    \vspace{-2mm}
\end{figure*}

\begin{figure}[!t]
    \centering
    \includegraphics[width=0.7\linewidth]{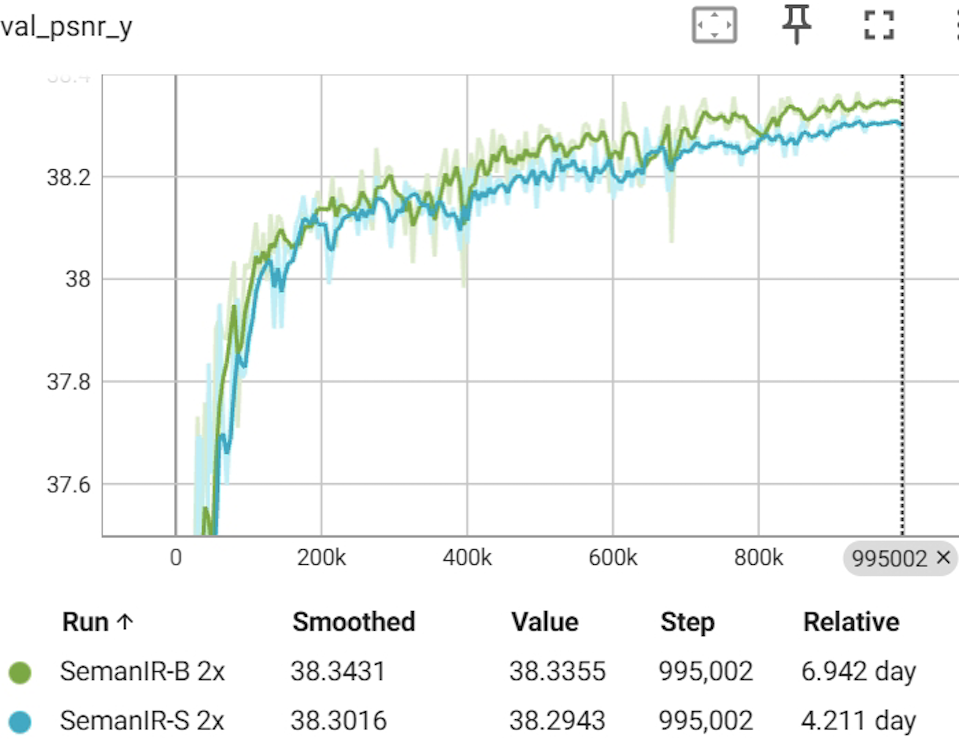}
    \vspace{-1mm}
    \caption{Training log shows the convergence of the proposed SemanIR during training. The upscaling factor is 2$\times$.}
    \label{fig:training_log}
    \vspace{-2mm}
\end{figure}

\section{Limitations}
\label{sec:appx_limit}
This study faces a task-specific limitation: each image restoration task requires training a separate network. While efforts have been made to train models for varying degradation levels within specific types, such as image denoising and removal of JPEG compression artifacts, this approach still leads to inefficiencies in model training. It constrains the utility of the trained networks. A potential future enhancement involves developing a mechanism enabling a network to handle diverse image degradation types and levels. Another challenge is the substantial parameter requirement of the proposed SemanIR, which operates within a tens-of-millions parameter budget. 
Deploying such a large image IR network on handheld devices with limited resources is challenging, if not unfeasible. 
Therefore, a promising research direction is the creation of more efficient versions of SemanIR, integrating non-local context more effectively, to overcome these limitations.

%-----------------------%
\section{Impact Statement}
\label{sec:appx_impact}
%-----------------------%
This paper introduces a transformer-based approach that significantly enhances the efficiency and performance of image restoration tasks, including image deblurring, JPEG CAR, image denoising, IR in adverse weather conditions, demosaicking, and image super-resolution. The proposed SemanIR's notable efficiency improvement holds promise for resource-effective implementations in real-world applications. 
This elevated performance creates opportunities for enhanced image quality across diverse domains. While our primary contribution lies in the technical aspects of Machine Learning, we are cognizant of potential societal impacts, particularly in healthcare, surveillance, and digital imaging. 
As with any technology, ongoing vigilance in ethical considerations during deployment is essential, ensuring responsible use and proactively addressing any unintended consequences.

\section{More Ablation Analyses}
\label{subsec:appx_moreabl}
Besides the ablation studies presented in our main manuscript, we further provide the following two analyses:
\begin{figure*}[!t]
    \centering
    \includegraphics[width=1.0\linewidth]{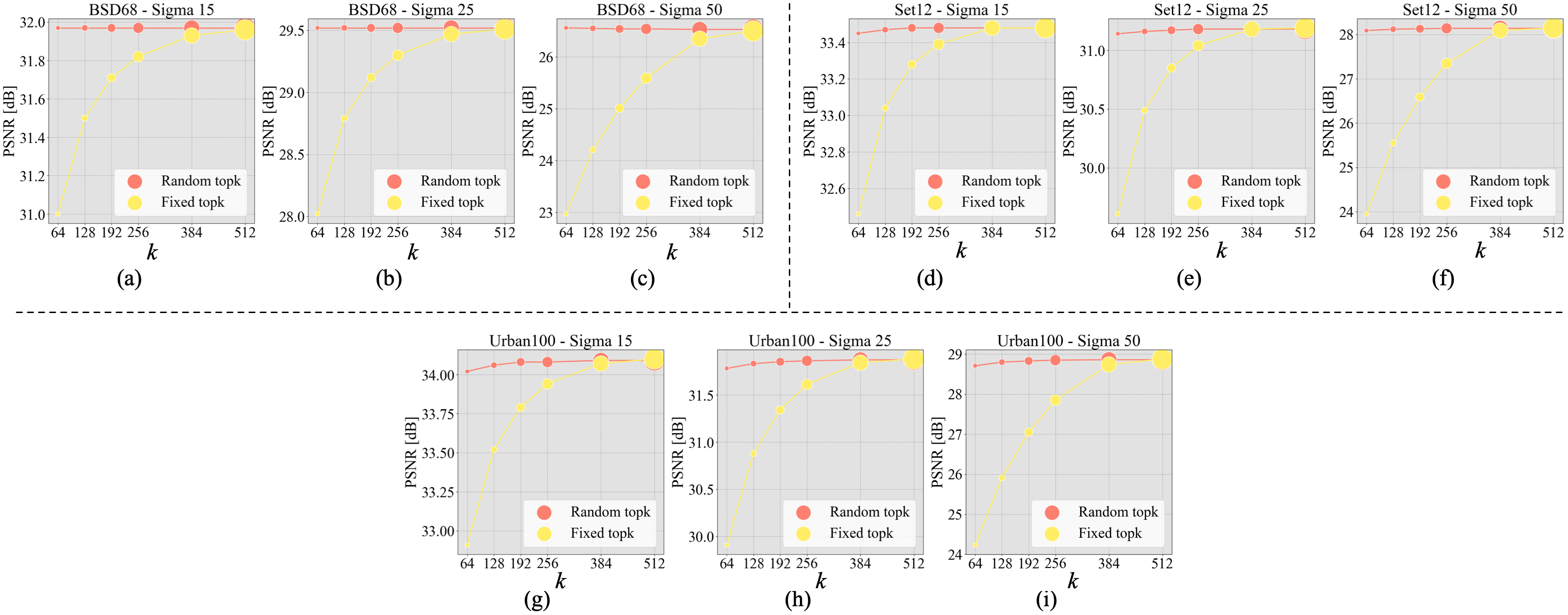}
    \vspace{-3mm}
    \caption{Ablation study on the impact of $k$ for \textbf{Grayscale Image Denoising} on BSD68 (a-c), Set12 (d-f), and Urban100 (g-i) datasets with $\sigma$ = \{15, 25, 50\}.}
    \label{fig:supple_fig_K_dn_gray}
    \vspace{-2mm}
\end{figure*}

\noindent\textbf{Convergence Visualization.} The training log of the proposed SemanIR for image SR is shown in Fig.~\ref{fig:training_log}. The log is reported for the PSNR on the Set5 dataset during training. Two versions of the proposed method including SemanIR-S and SemanIR-B are shown in this figure. As shown in this figure, the proposed network converges gradually during the training.

\noindent\textbf{The Impact of the $k$ in Key-Semantic Dictionary Construction under Various IR Tasks.} To explore how the $k$ value of top-k will affect the IR performance of the proposed SemanIR. We conduct exhaustive experiments on 
JPEG compression artifact reduction for color images under different QF values (\ie, QF = 
[10, 20, 30, 40]), 
image denoising for both color and grayscale images under different noise levels (\ie, $\sigma$ = [15, 25, 50]), as well as 
image SR under different scales (\ie, 2$\times$, 3$\times$, 4$\times$) with the proposed SemanIR. 
Note that all the experiments for each IR task are conducted under two kinds of top-k settings, \ie, (i) $k$ was randomly sampled from the range [64, 512] during the overall training phase, and (ii) $k$ was held constant at 512 throughout the training phase. For inference, $k$ was configured to the specified value for both settings.

The results of the JPEG CAR in terms of the hyper-parameters $k$ under different training settings during inference for color image are shown in 
Fig.~\ref{fig:supple_fig_K_jpeg_color}.
It is clear that for the color JPEG CAR task when $k$ is set to 64 during inference, there is a huge performance cat between the random top-k setting and the fixed top-k setting. In addition, the fixed top-k setting performs well or sometimes even a bit better than the random top-k setting only when $k$ is also set to the same number (\ie, 512). 
With the decrease of $k$ during inference for the fixed top-k setting, the PSNR drops largely marginally for all the datasets under every kind of degraded QF factor.

\begin{figure*}[!t]
    \centering
    \includegraphics[width=1.0\linewidth]{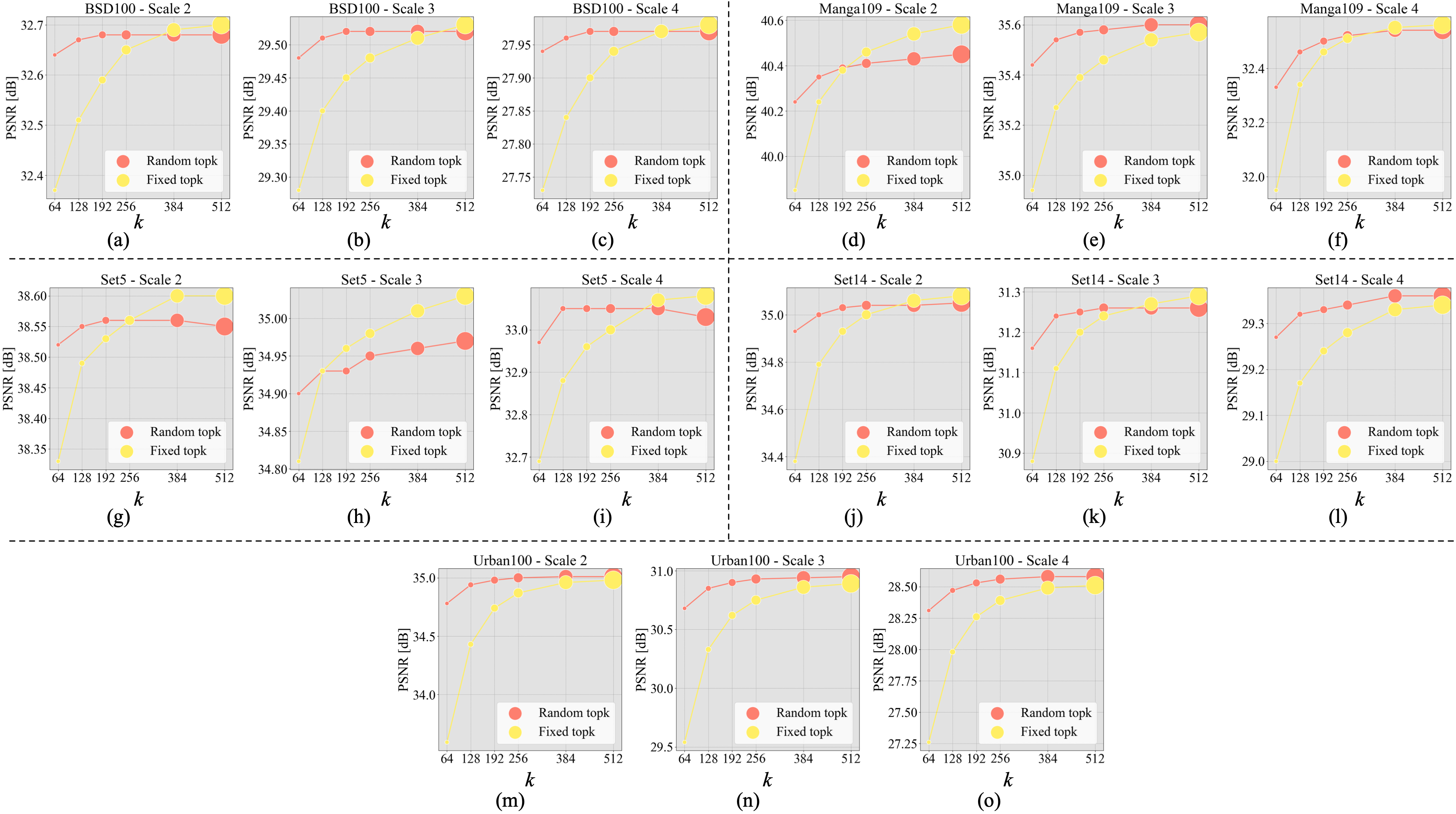}
    \vspace{-3mm}
    \caption{Ablation study on the impact of $k$ for \textbf{Image SR} with \textbf{SemanIR-B} on BSD100 (a-c), Manga109 (d-f), Set5 (g-i), Set14 (i-l) and Urban100 (m-o) datasets with $scale$ = (2$\times$, 3$\times$, and 4$\times$).}
    % {\{}2 $times$, 3$times$, x$times${\}}.}
    \label{fig:supple_fig_K_sr_base}
    \vspace{-2mm}
\end{figure*}

The results of the image denoising in terms of hyper-parameters $k$ under different training settings during inference for both color image and grayscale image are shown in Fig.~\ref{fig:supple_fig_K_dn_color} and Fig.~\ref{fig:supple_fig_K_dn_gray}. All the experimental results on various datasets (\ie, BSD68/CBSD68, Kodak24, McMaster, and Urban100) share a similar trend for image denoising compared to the JPEG CAR task. The random top-k setting can maintain a relatively stable PSNR score under different $k$ during inference compared to its fixed counterpart. In addition, a decent result can be obtained for the fixed top-k setting only when the $k$ is set to the same (\ie, 512) during the inference. 

The results of the image SR in terms of hyper-parameters $k$ under different training settings during inference for color images with different scale factors (\ie, 2$\times$, 3$\times$, and 4$\times$) are also provided in Fig.~\ref{fig:supple_fig_K_sr_base}. All experiments are conducted in various datasets (\ie, BSD100, Manga109, Set5, Set14, and Urban100). It shows that for datasets like BSD100, Set14, and Urban100, a similar trend can be also observed in Fig.~\ref{fig:supple_fig_K_sr_base} compared to JPEG CAR and image denoising tasks, \ie, the random top-k setting performs more stable regardless the change of the K during the inference. However, for Manga109 and Set5 dataset. The best PSNR is obtained by the fixed top-k setting (in (d) - (i) in Fig.~\ref{fig:supple_fig_K_sr_base}).

In general, based on all the experimental results mentioned above, we conclude that (1) the random top-k setting performs better than the fixed $k$ setting, and usually outperforms the latter by a large margin when $k$ is fixed to small values (\emph{i.e.}, 64, 128, 192, or 256.). (2) For the fixed top-k setting, if $k$ is set to big enough (\ie, 512) during inference, the fixed top-k setting can also achieve comparative performance or even better performance compared to the random top-k setting for several experiments (\eg color JPEG CAR in Fig.~\ref{fig:supple_fig_K_jpeg_color} (g) and Fig.~\ref{fig:supple_fig_K_jpeg_color} (k)). However, it is not always possible that the large fixed $k$ setting can be generalized to limited computation resources, and the model trained with large fixed $k$ usually needs the same $k$ for inference to maintain the performance, which leads to heavy computation resources needed even for inference.

To this end, we propose to decouple the way to use $k$ between training and inference. \ie, we can use the random sample $k$ during training while an optional fixed $k$ during inference without degenerating the overall performance. It makes it possible to deploy models that heavily rely on large GPU memory during training but to limited GPU resources while maintaining reliable performance during inference. This is also consistent with the way we implement the proposed attention block (\ie, we adopt a \textit{Torch-Mask} version that requires affordable large GPU memories during training compared to \textit{Torch-Gather} while adopting the \textit{Triton} version during inference) of SemanIR.

In addition, setting a predetermined $k$ value for each patch/pixel enhances computational efficiency. A fixed $k$ value facilitates parallel computation, particularly in attention operations. Conversely, making $k$ values learnable for each patch or pixel would significantly increase the complexity of the attention operation. Nonetheless, exploring the potential of learnable $k$ values for each patch or pixel represents an intriguing avenue for further investigation.

\begin{figure*}[!t]
    \vspace{1mm}
    \centering
    \includegraphics[width=1.0\linewidth]{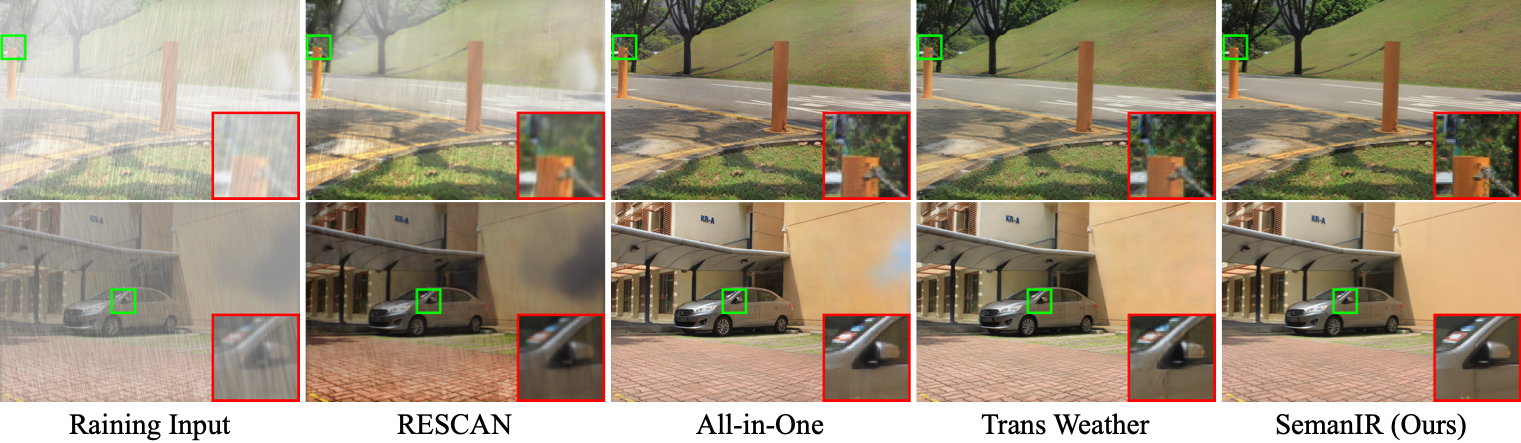}
    \caption{Visual comparison for restoring images in AWC. Best viewed by zooming.}
    \label{fig:awc_x4}
    % \vspace{-0.5cm}
\end{figure*}

\section{More visual Results}
\label{sec:appx_results}
To further support the effectiveness of the proposed SemanIR intuitively. We provide more visual comparison in terms of image deblurring, JPEG CAR, image denoising, and image SR below.

\begin{figure*}[!t]
    \centering
    \includegraphics[width=1.0\linewidth]{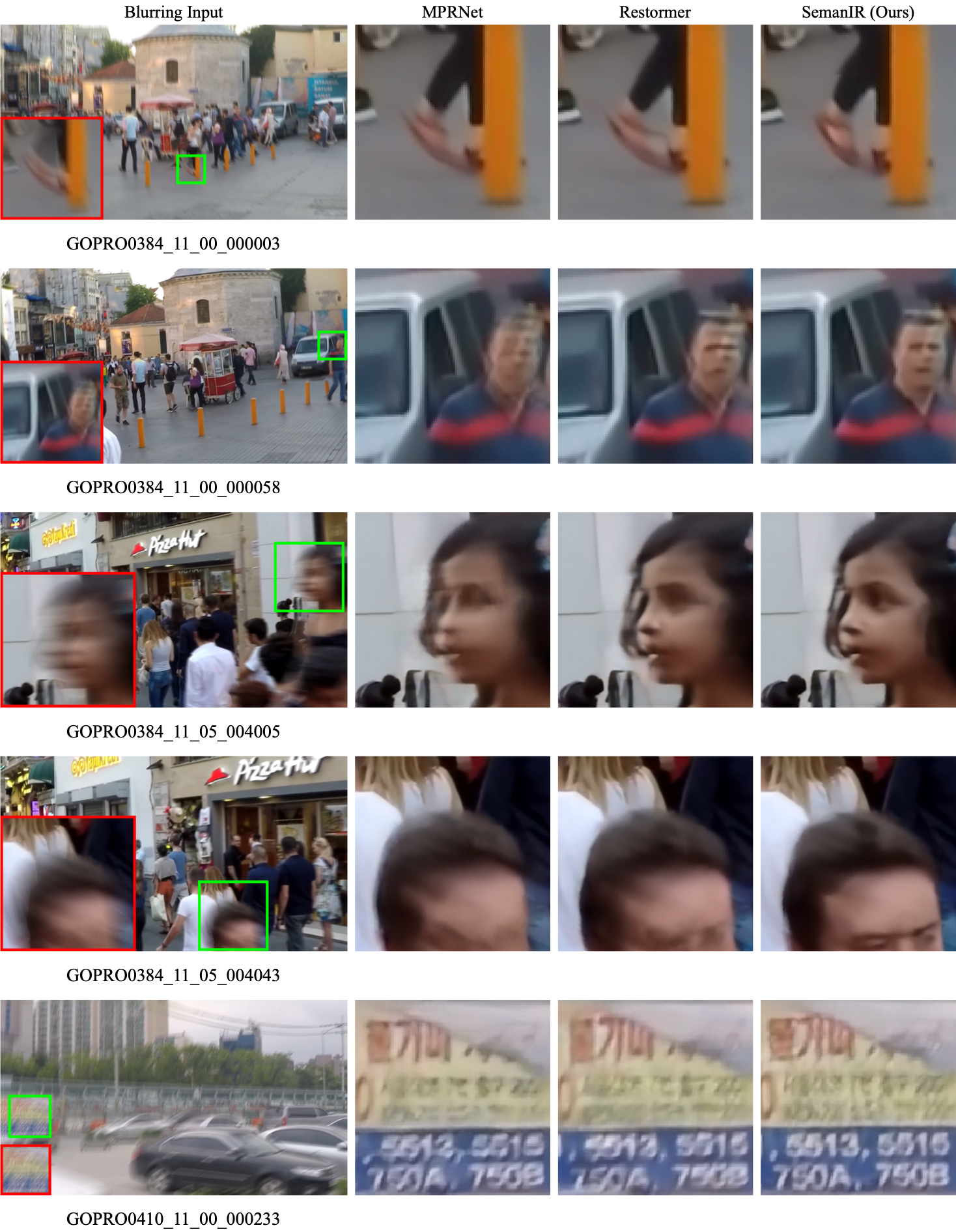}
    \caption{Visual comparison with single image motion deblurring on GoPro dataset. Best viewed by zooming.}
    \label{fig:supple_db_gopro}
\end{figure*}

\noindent\textbf{Image Deblurring}
The visual results for single image motion deblurring are shown in Fig.~\ref{fig:supple_db_gopro}. As shown in this figure, the proposed method can effectively remove the motion blur in the input images and restore more details such as the facial contour, and the characters compared to MPRNet or Restormer.

\begin{figure}[!t]
    \centering
    \includegraphics[width=0.93\linewidth]{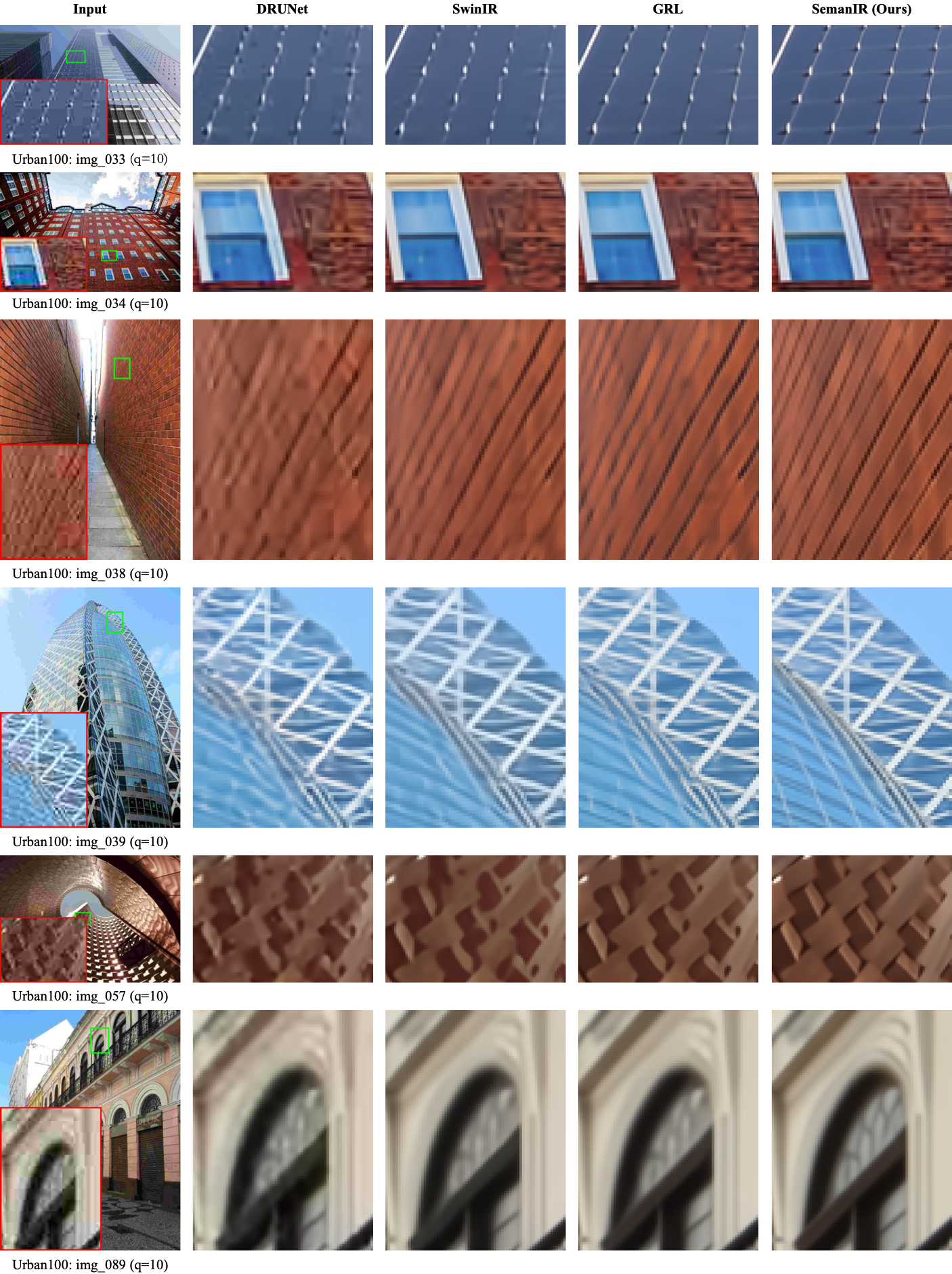}
    \caption{Visual comparison of color JPEG CAR on Urban100 dataset. Best viewed by zooming.}
    \label{fig:visual_jpeg_color_urban}
\end{figure}

\noindent\textbf{JPEG Compression Artifact Removal}
For JPEG compression artifact removal, the visual results for color images on the Urban100 dataset are shown in Fig.~\ref{fig:visual_jpeg_color_urban}. The proposed method achieves state-of-the-art performance in removing the blocking artifacts in the input images.

\begin{figure*}[!t]
    \centering
    \includegraphics[width=0.93\linewidth]{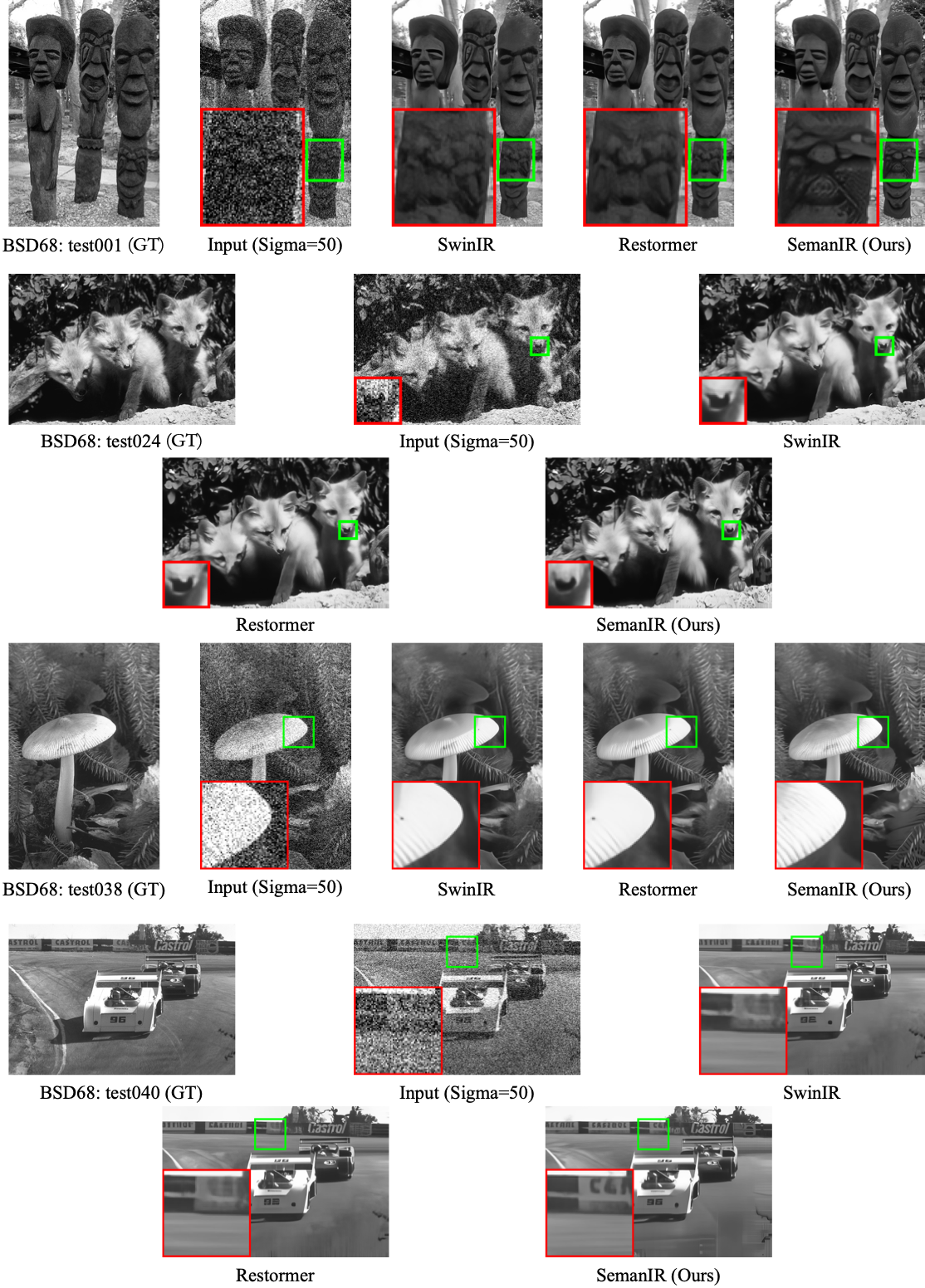}
    \caption{Visual comparison with image denoising on BSD68 dataset. Best viewed by zooming.}
    \label{fig:supple_dn_bsd68}
\end{figure*}

\begin{figure*}[!t]
    \centering
    \includegraphics[width=0.93\linewidth]{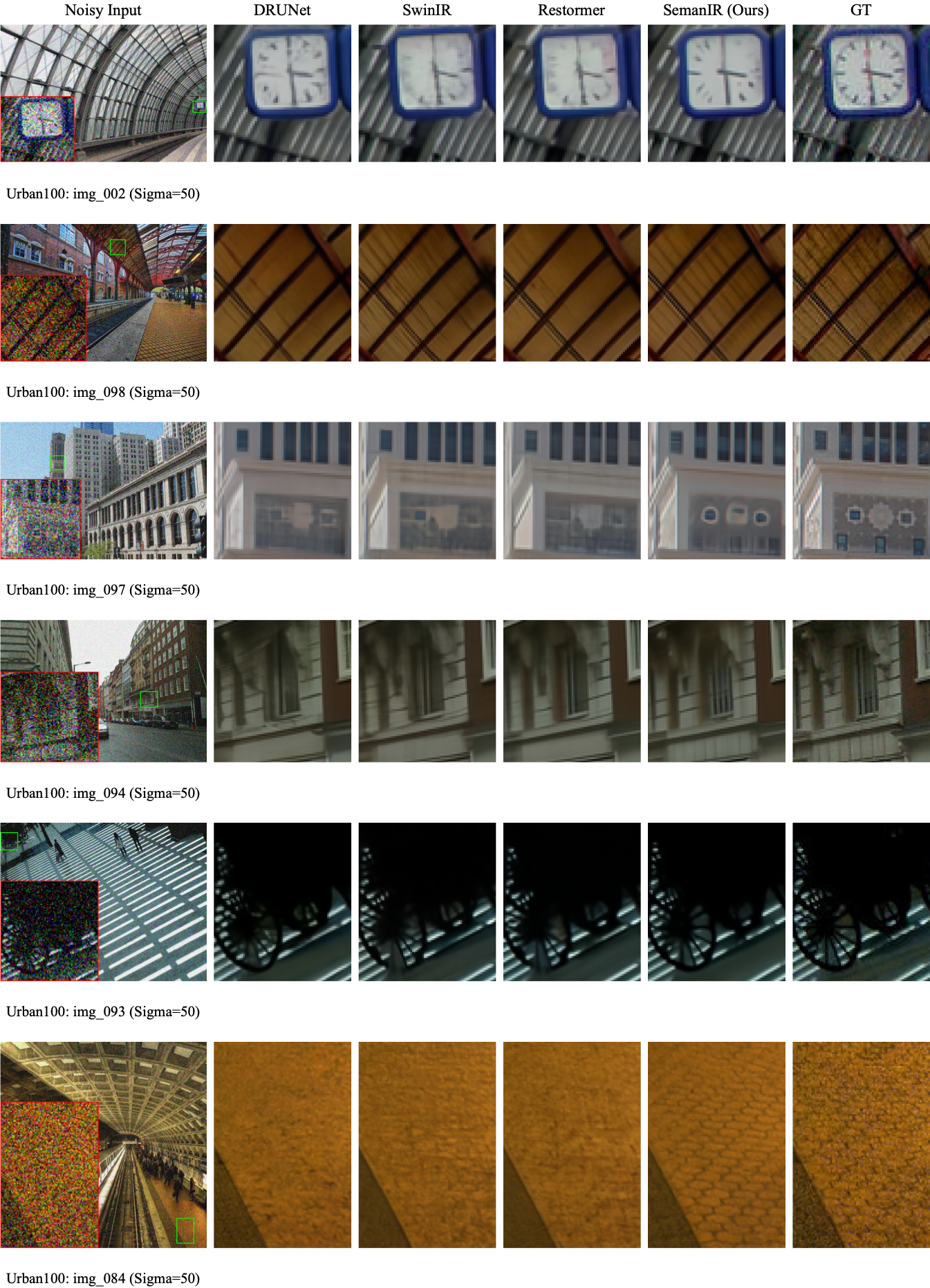}
    \caption{Visual comparison with image denoising on Urban100 dataset. Best viewed by zooming.}
    \label{fig:supple_dn_urban100}
\end{figure*}

\noindent\textbf{Image Denoising}
The qualitative results for image denoising on the BSD68 and the Urban100 dataset are shown in Fig.~\ref{fig:supple_dn_bsd68} (grayscale image) and Fig.~\ref{fig:supple_dn_urban100} (color image). It is clear that for both the grayscale and color inputs, the proposed SemanIR can remove the noise in the noisy input images and recover more realistic textural details in the restored images.

\noindent\textbf{IR Adverse Weather Conditions.}
The qualitative results for IR in AWC on the Test1~\cite{li2020all, li2019heavy} dataset are shown in Fig.~\ref{fig:awc_x4}. It shows a challenging case but our method can restore better structural content and clearer details.

\begin{figure*}[!t]
    \centering
    \includegraphics[width=0.85\linewidth]{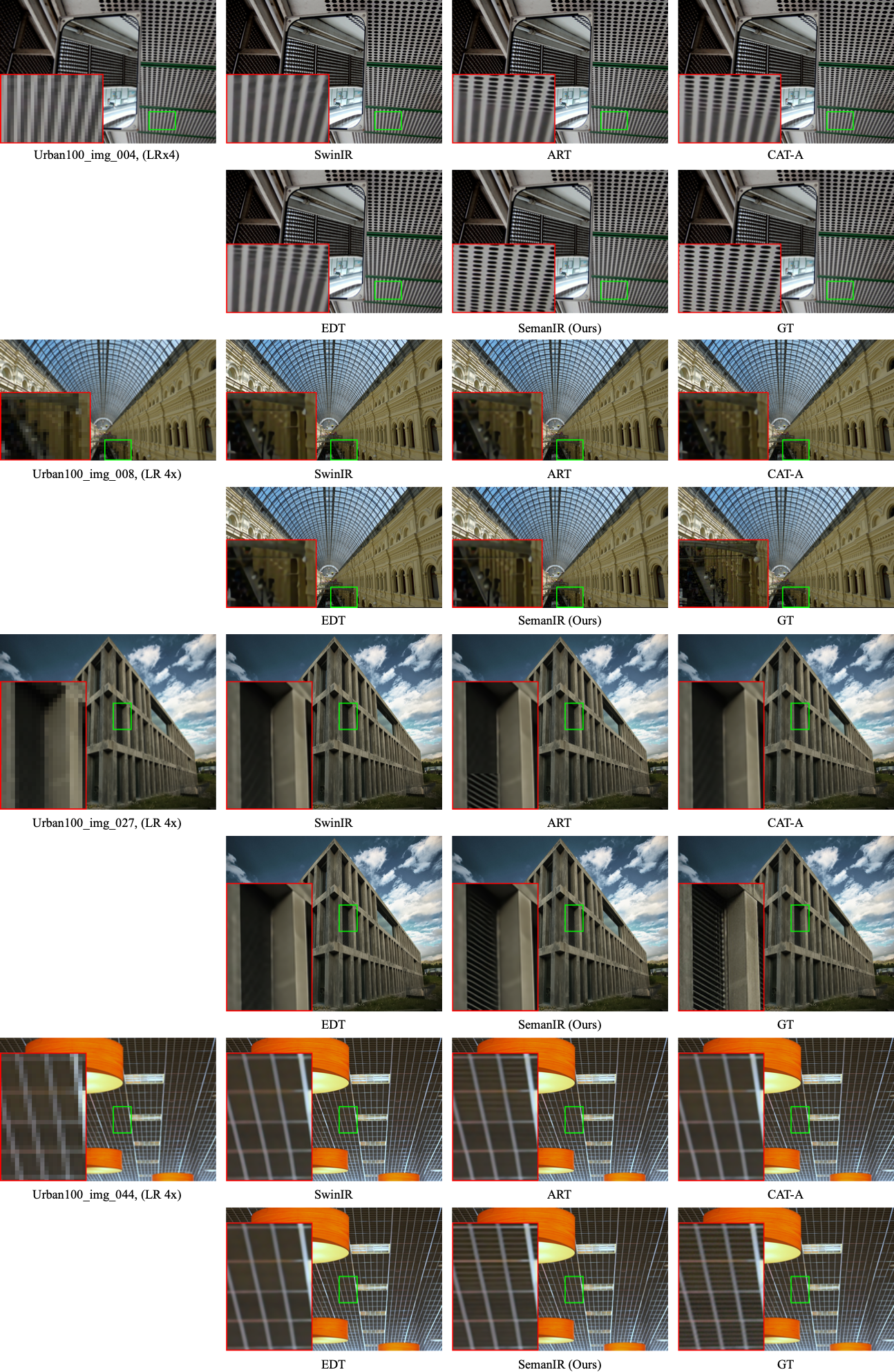}
    \caption{Visual comparison (4$\times$) with image SR on Urban100 dataset. Best viewed by zooming.}
    \label{fig:supple_sr_x4_urban}
\end{figure*}

\begin{figure*}[!t]
    \centering
    \includegraphics[width=0.95\linewidth]{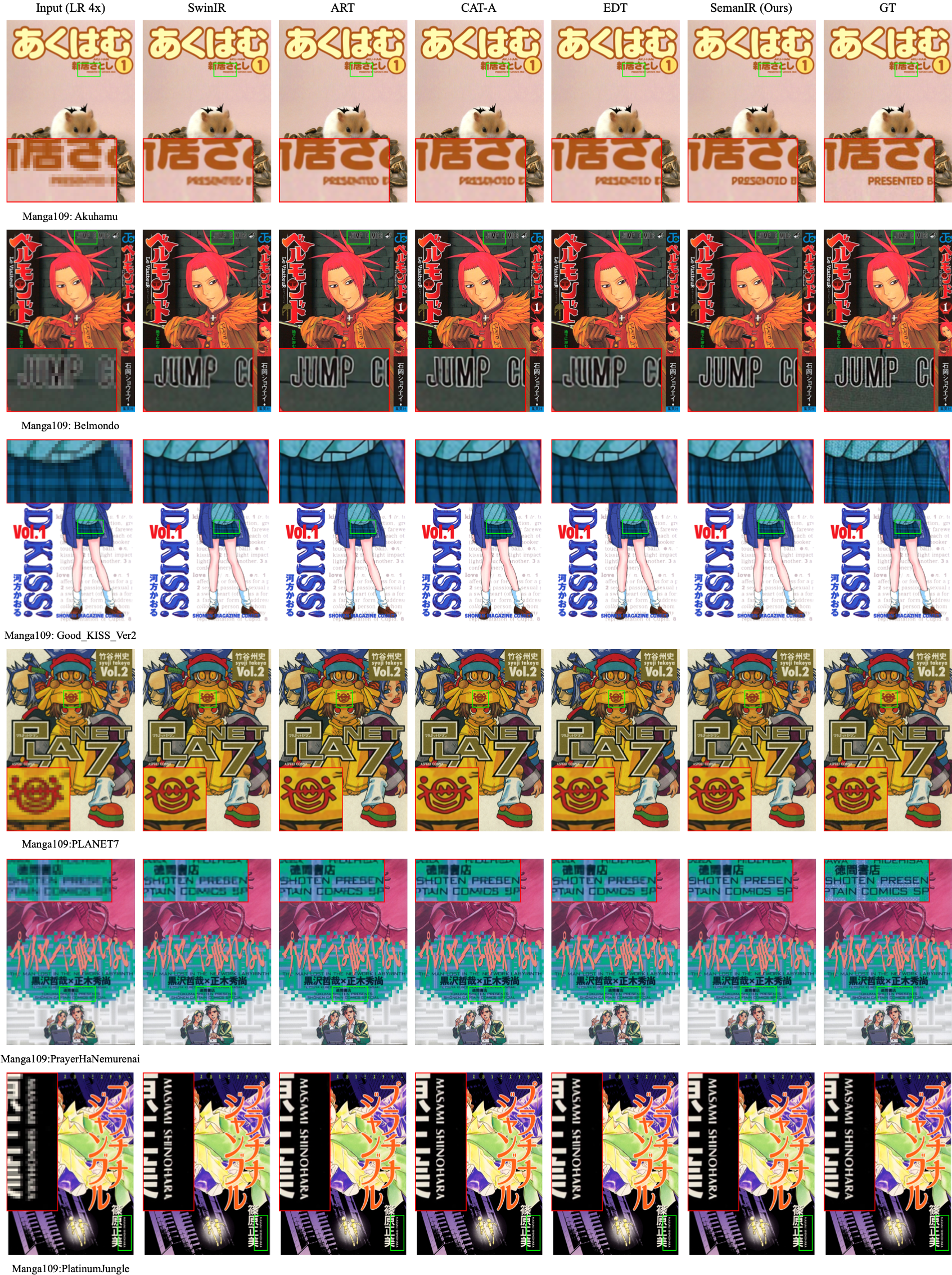}
    \caption{Visual comparison (4$\times$) with image SR on Manga109 dataset. Best viewed by zooming.}
    \label{fig:supple_sr_x4_manga}
\end{figure*}

\noindent\textbf{Image SR.}
The comparison of visual results of different image SR methods is shown in 
Fig.~\ref{fig:supple_sr_x4_urban} and Fig.~\ref{fig:supple_sr_x4_manga}.
Fig.~\ref{fig:supple_sr_x4_urban} shows the results on the Urban100 dataset, and Fig.~\ref{fig:supple_sr_x4_manga} shows the results on the Manga109 dataset. The proposed SemanIR can restore more missing details in the LR images compared to other state-of-the-art methods like SwinIR, ART, CAT, and EDT.

\end{document}